\documentclass[10pt,twocolumn,letterpaper]{article}

\usepackage{iccv}
\usepackage{times}
\usepackage{epsfig}
\usepackage{graphicx}
\usepackage{amsmath}
\usepackage{amssymb}
\usepackage{multirow} %
\usepackage{microtype} %
\usepackage{booktabs} %
\usepackage{caption,subcaption} %
\usepackage{textcomp}%
\usepackage[usenames,dvipsnames]{color}%
\usepackage{colortbl}%

\newcommand*\rot{\rotatebox{90}}

\definecolor{mygray}{gray}{.9}

\usepackage[pagebackref=true,breaklinks=true,letterpaper=true,colorlinks,bookmarks=false]{hyperref}

 \iccvfinalcopy 

\def\iccvPaperID{1197} 

\ificcvfinal\fi
\begin{document}

\title{Significance-aware Information Bottleneck for Domain Adaptive \\ Semantic Segmentation}

\author{Yawei Luo$^{1,2}$,\hspace{2mm} Ping Liu$^{2}$,\hspace{2mm} Tao Guan$^{1}$,\hspace{2mm} Junqing Yu$^{1}$, \hspace{2mm} Yi Yang$^{2}$ \vspace{0.5cm} \\ 
$^1$School of Computer Science \& Technology, HUST ~ $^2$CAI, UTS
}

\maketitle

\begin{abstract}
For unsupervised domain adaptation problems, the strategy of aligning the two domains in latent feature space through adversarial learning has achieved much progress in image classification, but usually fails in semantic segmentation tasks in which the latent representations are overcomplex. In this work, we equip the adversarial network with a ``significance-aware information bottleneck (SIB)'', to address the above problem. The new network structure, called SIBAN, enables a significance-aware feature purification before the adversarial adaptation, which eases the feature alignment and stabilizes the adversarial training course. In two domain adaptation tasks, i.e., GTA5 $\rightarrow$ Cityscapes and SYNTHIA $\rightarrow$ Cityscapes, we validate that the proposed method can yield leading results compared with other feature-space alternatives. Moreover, SIBAN can even match the state-of-the-art output-space methods in segmentation accuracy, while the latter are often considered to be better choices for domain adaptive segmentation task.

\vspace{-0.5cm}
\end{abstract}

\section{Introduction}
Semantic segmentation aims to assign each image pixel a category label. The recent adoption of Convolutional Neural Networks (CNNs) yields various of best-performing methods~\cite{long2015fcn,chen2018deeplab,luo2018macro} for this task, but the achievement is at the price of a huge amount of dense pixel-level annotations obtained by expensive human labor. An alternative would be resorting to simulated data, such as computer-generated scenes~\cite{richter2016gta5,ros2016synthia}, which can make unlimited amounts of labels available. However, models trained with the simulated images, no matter how perfect they perform on the simulation environment, fail to achieve a same or even similar satisfactory performance on realistic images. The reason behind this performance drop lies in the different data distributions of the two domains, typically known as domain shift~\cite{shimodaira2000improving}. 

\begin{figure}
\centering
\includegraphics[width=0.9\linewidth]{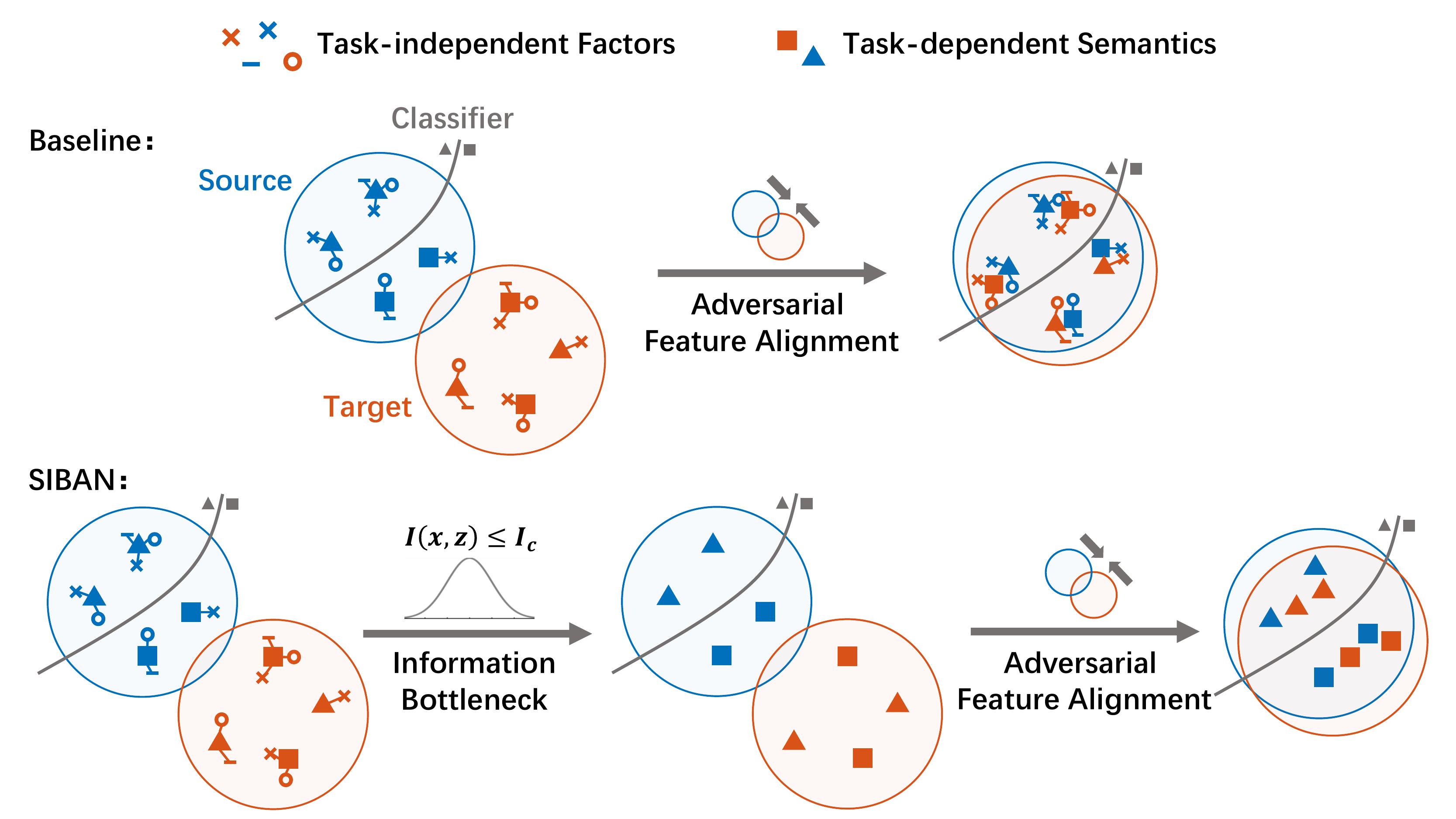}
\caption{Comparison of the baseline method and SIBAN. The baseline method aligns the latent features directly. As the crude features contain various of task-independent factors, these features are prone to be wrongly aligned between two domains. SIBAN addresses this issue by employing an information bottleneck before the adversarial feature adaptation. The information bottleneck filters out the nuisance factors and maintains pure semantic information. Since the two domains essentially overlap in semantic-level, such purified features can facilitate the following alignment and stabilize the adversarial training course.}
\label{fig:Intro}
\vspace{-0.5cm}
\end{figure}

\textit{Domain Adaptation (DA)} approaches~\cite{saito2017maximum,tsai2018OutputSpace,hoffman2017cycada} are proposed to bridge the gap between the source and target domains. These methods tend to align the two domains in latent feature space so that a classifier trained on source data can also be applied to target samples. Despite the fact that great success has been made on the image level classification task~\cite{long2015MMD,tzeng2017adversarial,motiian2017unified}, applying the latent space adaptation to semantic segmentation is non-trivial. The reasons are summarized as twofold. On the one hand, latent space adaptation for semantic segmentation may suffer from the complexity of high-dimensional features which encode various visual cues: appearance, shape and context, \emph{etc}. Some of the task-independent nuisance factors might be easily involved in the encoded representation and mislead the domain alignment. On the other hand, in the adversarial domain adaptation framework~\cite{goodfellow2014gan} which becomes popular in this field, the redundant information from the task-irrelevant factors might give excessive cues to the discriminator. The excessive cues lead the discriminator ``unnecessary"  high accuracy at the wrong time and produce uninformative gradients. All of this, unfortunately,  will make the adversarial training process unstable, as pointed out in~\cite{luo2018macro,karras2018progressive,peng2019variational}. 

Being hampered by the difficulties in feature-space adaptation, the current tendency turns to explore the DA possibility in other spaces, including pixel (input) space and segmentation (output) space. The common idea of the pixel-space adaptations is to force the input images to look like from the same domain, thus decreasing the domain shift from the headstream. While segmentation-space approaches are based on the observation that the segmentation results usually share a significant amount of similarities on the spatial layout and local context. Currently, these two lines of work have produced leading results on the semantic segmentation task while the feature-space adaptation appears eclipsed in front of them. 
Taking the DA task GTA5~\cite{richter2016gta5} $\rightarrow$ Cityscapes~\cite{cordts2016cityscapes} as an example, there is a big difference in segmentation accuracy between the feature-space and output-space adaptation method: $29.2\%~vs~34.8\%$~\cite{hoffman2017cycada} on VGG-16~\cite{long2015fcn},  $31.7\%~vs~37.0\%$~\cite{hoffman2017cycada} on DRN-26~\cite{yu2017dilated}, and $39.3\%~vs~41.4\%$~\cite{tsai2018OutputSpace} on ResNet-101~\cite{he2016resnet}, respectively. 
The performance gap is so significant that it is justifiable the previous methods choose output-space adaptation as their first choice.


Now a question arises: is the feature-space adaptation really infeasible for the semantic segmentation task? This paper gives a negative answer. As previously analyzed, the obstacles in feature-space adaptation consist in 1) the difficulty of aligning the complicated latent representations between two domains and 2) the difficulty of training the adversarial network stably because of the overly accuracy of the discriminator. Accordingly, we propose Significance-aware Information Bottlenecked Adversarial Network (\textbf{SIBAN}), which overcomes the two obstacles above. 

Our approach is inspired by the \textit{information bottleneck (IB)} theory~\cite{tishby1999information}, where the learned latent representation $Z$ needs to make a consistent prediction with the ground-truth labels $Y$ while simultaneously contains the least mutual information $I(X,Z)$ with the given input $X$. In our framework, the information bottleneck is employed to compact the complicated latent representations to facilitate the feature alignment and adversarial training.

On the one hand, by enforcing a constraint on the mutual information $I(X,Z)$, we encourage the feature extractor to filter out those task-independent nuisance factors while only keeping the task-dependent factors. In our semantic segmentation task, the task-dependent factor corresponds to the pure semantic information. Since in our \emph{simulated $\rightarrow$ real} setting, the two domains vary a lot at \emph{visual level}, but overlap at \emph{semantic level}, such pure semantic information is usually domain-invariant. On the other hand, in the adversarial learning-based framework for adaptation, utilizing the information bottleneck prevents $D$ from the distractions introduced by task-irrelevant factors, which is difficult for the vanilla generator $G$ to depress. As a matter of fact, our proposed network effectively modulates the $D$'s behavior, thus can stabilize the adversarial training process.

Unlike the \emph{implicit} way for domain-invariant feature detection utilized by the vanilla adversarial learning framework~\cite{hoffman2017cycada,tsai2018OutputSpace},  the information bottleneck chooses an \emph{explicitly} constraint to filter out the unwanted information during the adaptation. And therefore, introducing an IB into adversarial learning framework combines the respective advantages in this aspect and makes distilling domain-invariant information between different domains more effectively and efficiently, which is proven in the experiment of this work. 

Moreover, to deal with the long-tailed data distribution problem~\cite{tan2017distant} introduced by the unbalanced pixel number between different classes, we propose a novel layer, which is named ``Significance Aware Layer".  By introducing this layer into the IB module, our framework takes the channel-wise significance of each semantic feature into consideration and keeps balanced information constraints between them based on their respective significance. We call this newly designed module as Significance-aware Information Bottleneck (\textbf{SIB}), the whole framework as Significance-aware Information Bottlenecked Adversarial Network (SIBAN).

On the whole, our contributions are summarized below.
\begin{itemize}
\item We propose a significance-aware information bottlenecked adversarial network (SIBAN) for feature-space domain adaptive semantic segmentation, which combines the advantages from Information Bottleneck theory and Adversarial Learning framework respectively.  To our knowledge, this is the first time to successfully utilize information bottleneck strategy for this challenging, dense labeling task.

\item We propose a Significance-aware IB (SIB) module and integrate it into our framework.  By taking advantage of this module, our framework is able to balance the information constraint between different classes, for maintaining the final performance on the classes which are rare among datasets.

\item We theoretically and experimentally prove the effectiveness of our approach, which achieves the leading adaptation result in feature space and performs on par with the state-of-the-art input/output-space adaptations.

\end{itemize} 

\section{Related Work}
\subsection{Domain Adaptive Semantic Segmentation}
Ben-David \emph{et al.}~\cite{ben2010theory} have proven that the adaptation loss is bounded by three terms, \emph{e.g.}, the expected loss on source domain, the domain divergence, and the shared error of the ideal joint hypothesis on the source and target domain. Because the first term corresponds to the well-studied supervised learning problems and the third term is considered sufficiently low, the majority of recent works lay emphasis on the second term. In this spirit, some approaches focus on the distribution shift in the latent feature space \cite{solomon2015WGAN,hoffman2016fcns,liu2016coupled,kim2017relations,tzeng2017adversarial,sankaranarayanan2017unsupervised}. Nevertheless, most of such methods only achieve in classification task while failing in segmentation. With a few exceptions, Hoffman \emph{et al.}~\cite{hoffman2016fcns} employed adversarial network to align the feature representations between domains and additionally appended category statistic constraints to the adversarial model. Apart from the feature-space DA, some methods address the problem in the pixel space \cite{li2018Grad-GAN,bousmalis2017pixelDA}, which relates to the style transfer approaches~\cite{zhu2017cycle,choi2017stargan} to make images indistinguishable across domains. Joint consideration of pixel- and feature-space domain adaptation is studied in~\cite{hoffman2017cycada}. For segmentation task, it is also found that aligning the segmentation space is a more effective DA strategy~\cite{tsai2018OutputSpace,chen2018learning}. Besides the adversarial training-based DA methods~\cite{hoffman2017cycada,tsai2018OutputSpace,li2018Grad-GAN}, other lines of work on semantic segmentation borrow the idea from self-training~\cite{rosenberg2005semi} or co-training~\cite{zhou2005cotrain}. The self-training-based DA~\cite{saito2017asymmetric,zou2018unsupervised} attempts to assign pseudo labels to target images and then use these labels to train the target model directly. While the co-training-based DA~\cite{saito2017maximum,luo2018taking} aims to detect the domain-invariant features by maximizing the consensus of the multiple classifiers.

\subsection{Information Bottleneck}
Information bottleneck~\cite{tishby1999information} (IB) tends to enforce an upper bound on the mutual information $I(X,Z)$ between the latent representation $Z$ learned by the encoder and the original input $X$. As pointed in~\cite{tishby1999information}, for a supervised learning task, IB encourages $Z$ to be predictive of the label $Y$, and simultaneously, push the $Z$ to ``forget” the original input $X$ as much as possible. This is equivalent to upperbound a $Kullback-Leibler (KL-)$ divergence between the joint probability $P(X,Z)$ and the product of the marginals $P(X) \times P(Z)$ to a specific bottleneck value $I_c$. Although the information bottleneck principle is appealing, it suffers from the fact that mutual information computation is computationally challenging~\cite{slonim2002information}, which is especially hard to be instantiated in the context of CNNs. Inspired by a similar approach in variational autoencoders (VAE)~\cite{kingma2013auto}, recent methods~\cite{alexander2017variational,peng2019variational} implemented the IB in practical deep models by leveraging a variational bound and the reparameterization trick. This paper follows such strategy to instantiate the IB in the context of adversarial learning-based domain adaptation.

\begin{figure*}[ht]
\centering
\includegraphics[width=0.87\linewidth]{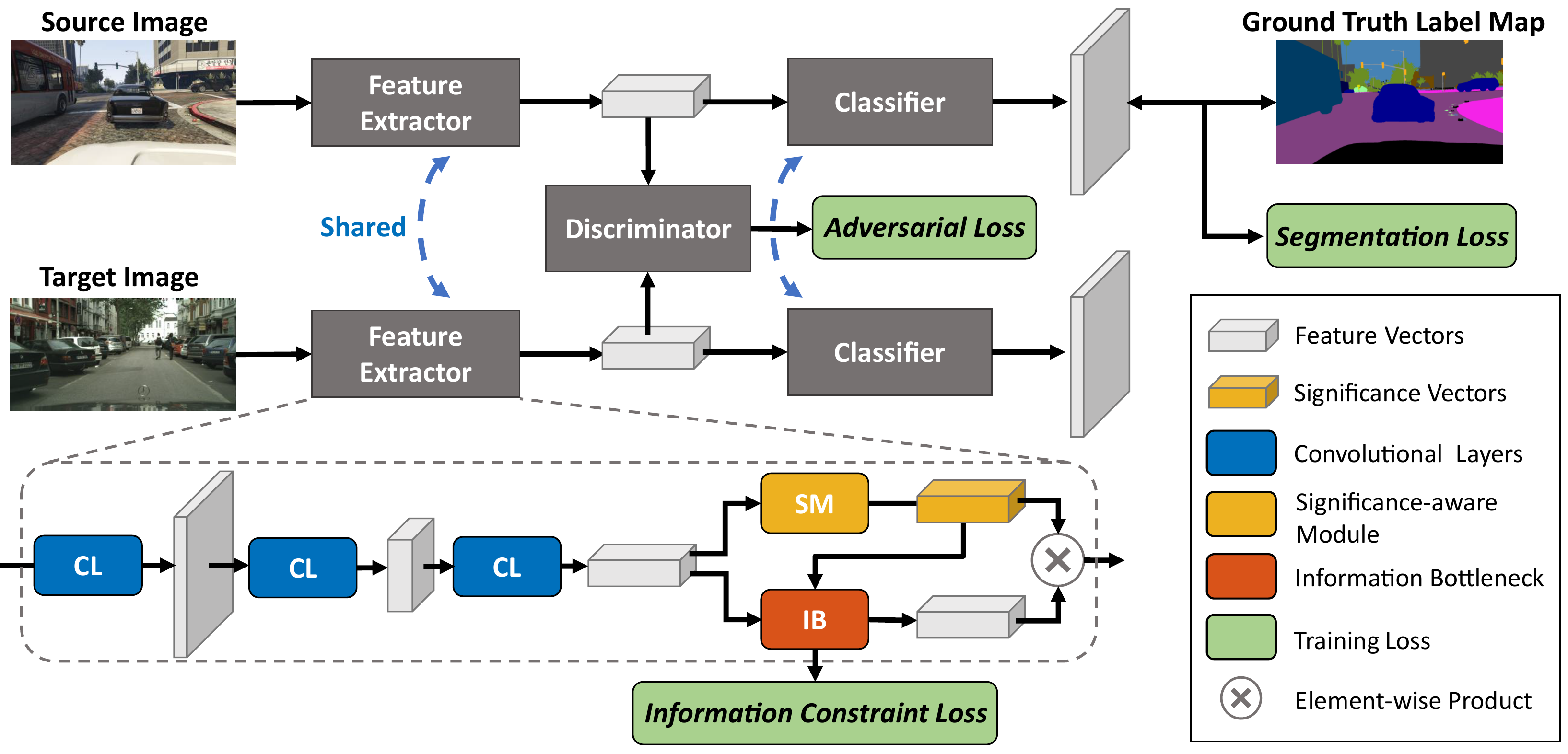}
\caption{The network architecture of the proposed SIBAN.}
\label{fig:main}
\vspace{-0.5cm}
\end{figure*}

\section{Method}
\subsection{Problem Settings and Overall Idea}
\label{Problem setting}
We focus on the problem of unsupervised domain adaptation (UDA) in semantic segmentation, where we have access to the labeled source dataset $\{\mathbf{x}^s_i, \mathbf{y}^s_i\}$ and the unlabeled target dataset $\{\mathbf{x}^t_i\}$. The goal is to learn a model $G$ that can correctly predict the pixel-level labels for the target data $\{\mathbf{x}^t_i\}$ by the information from $\{\mathbf{x}^s_i, \mathbf{y}^s_i\}$ and $\{\mathbf{x}^t_i\}$. To facilitate the discussion, we divide $G$ into a feature extractor $F$ and a classifier $C$, where $G = C \circ F$. Accordingly, we denote the latent representation $z$ as $z = F(\mathbf{x})$ and the final segmentation prediction as $\mathbf{\hat{y}} = C \circ F(\mathbf{x})$. 

Traditional feature-level adaptations~\cite{hoffman2016fcns,hoffman2017cycada,tsai2018OutputSpace} consider two aspects in dealing with the problem discussed above. First, these methods train a model $G$ to distill knowledge from labeled data by minimizing the task loss in the source domain, which is formalized as a supervised problem:
\vspace{-0.1cm}
\begin{equation}
\mathcal{L}_{seg} (F, C) = \mathbb{E}_{\mathbf{x},\mathbf{y} \sim p(\mathbf{x}^S, \mathbf{y}^S)}[ \ell (C \circ F(\mathbf{x}), \mathbf{y}) ] \; ,
\label{eq-basic}
\end{equation}
where $\mathbb{E}[ \cdot ]$ denotes statistical expectation and $\ell(\cdot, \cdot)$ is an appropriate loss function, such as multi-class cross entropy.

Second, during the training process, those feature-level adaptation methods also make $F$, the submodule in $G$, to learn domain-invariant features. Ideally, the domain-invariant features should confuse a domain discriminator $D$ which aims at distinguishing the features extracted between the source and target domains. This is achieved by minimaxing an adversarial loss:
\vspace{-0.1cm}
\begin{equation}
\begin{aligned}
\mathcal{L}_{adv} (F, D) = & -\mathbb{E}_{\mathbf{x} \sim p(\mathbf{x}^S)}[ \log (D(F(\mathbf{x})))] \\
& -\mathbb{E}_{\mathbf{x} \sim p(\mathbf{x}^T)}[ \log (1-D(F(\mathbf{x}))) ] \; .
\label{eq-basic-adversarial}
\end{aligned}
\end{equation}

However, as mentioned above, there is a significant limitation in previous feature-space adversarial learning methods~\cite{hoffman2016fcns,hoffman2017cycada,tsai2018OutputSpace}: there is no explicit constraint to prevent the network from encoding task-independent nuisance factors into the latent features, which makes the adaptation difficult and the adversarial training unstable. To handle the issue, we propose to distill the task-dependent parts from the crude features and conduct the adaptation based on these ``purified" representations, thus helping the feature adaptation and stabilizing the adversarial training.

\subsection{Information Constrained Domain Adaptation}\label{ICDA}
The pipeline of our network is shown in Fig.~\ref{fig:main} where we utilize a simple feature-space adversarial network as the backbone. To purify the encoded latent representation, we adopt an information constraint on the latent space, encouraging $F$ to encode only task-dependent semantic features into the representations. Built upon the recently developed information theories for deep learning~\cite{alexander2017variational,peng2019variational}, 
we achieve such constraint by employing a variational information bottleneck into the feature extractor $F$, which is shared among the source domain and target domain respectively. In this case, we obtain the following objective function:
\vspace{-0.1cm}
\begin{equation}
\begin{aligned}
F^*, C^*, D^* = & \arg\min_{F,C}\max_D \mathcal{L}_{seg} (F, C) + \lambda \mathcal{L}_{adv} (F, D) \\
s.t. \qquad & \mathbb{E}_{\mathbf{x} \sim p(\mathbf{x}^S)}(\mathrm{KL}[F(\mathbf{z}|\mathbf{x}) || r(\mathbf{z})]) \leq I_c, \\ 
& \mathbb{E}_{\mathbf{x} \sim p(\mathbf{x}^T)}(\mathrm{KL}[F(\mathbf{z}|\mathbf{x}) || r(\mathbf{z})]) \leq I_c.
\label{eq:constraint}
\end{aligned}
\end{equation}
where $r(\mathbf{z})$ denotes a prior marginal distribution of $\mathbf{z}$, which is modeled as a standard Gaussian $\mathcal{N}(0; I)$ in this paper. The intuitive meaning of variational IB is clear: the larger the $KL$-divergence between $F(\mathbf{z}|\mathbf{x})$ and $r(\mathbf{z})$, the stronger the dependence between $\mathbf{x}$ and $\mathbf{z}$, indicating that $\mathbf{z}$ encodes more information from $\mathbf{x}$, in which case some of them might be not task-related and therefore harmful to the adaptation. Therefore, by enforcing the $KL$- divergence to a threshold $I_c$ and minimizing the task loss, we can explicitly remove the task-independent factors from $\mathbf{z}$. 

We can equivalently optimize Eq.~\ref{eq:constraint} by introducing two Lagrange multipliers: $\beta^{S} \geq 0$ for the source domain, and $\beta^{T} \geq 0$ for the target domain:

\begin{equation}
\begin{aligned}
F^*, C^*, D^* = & \arg\min_{F,C}\max_D \mathcal{L}_{seg} (F, C) + \lambda \mathcal{L}_{adv} (F, D) \; + \\
&\beta^S(\mathbb{E}_{\mathbf{x} \sim p(\mathbf{x}^S)}(\mathrm{KL}[F(\mathbf{z}|\mathbf{x}) || r(\mathbf{z})]) - I_c) \; + \\
&\beta^T(\mathbb{E}_{\mathbf{x} \sim p(\mathbf{x}^T)}(\mathrm{KL}[F(\mathbf{z}|\mathbf{x}) || r(\mathbf{z})]) - I_c).
\label{eq:objective}
\end{aligned}
\end{equation}

To simplify the formulation, we define the last two items of Eq.~\ref{eq:objective} as the information constraint losses $\mathcal{L}_{ic}^{S}$ and $\mathcal{L}_{ic}^{T}$ for source and target domain, respectively. Accordingly, the overall training loss can be rewritten as:

\begin{equation}
\begin{aligned}
\mathcal{L}_{Overall}(F,C,D) = &\mathcal{L}_{seg} (F, C) + \lambda \mathcal{L}_{adv} (F, D) \; + \\
&\beta^S\mathcal{L}_{ic}^{S} (F) + \beta^T\mathcal{L}_{ic}^{T} (F).
\label{eq:overall loss}
\end{aligned}
\end{equation}

Unlike the traditional information bottleneck methods~\cite{chalk2016relevant,alexander2017variational} that uses a fixed $\beta$, we follow the suggestion of ~\cite{peng2019variational} to adaptively update $\beta^S/\beta^T$ via dual gradient descent. The motivation behind is intuitive: the more bias should be given to the feature purification when the encoded information overflow the bottleneck and vice versa, to enforce a specific constraint $I_c$ on the mutual information. Specifically, we train the network to minimax the overall loss $\mathcal{L}_{Overall}(F,C,D)$ by alternating between optimizing $F$, $C$, $D$, $\beta^S$ and $\beta^T$ until the loss converges.
\vspace{-0.1cm}
\begin{equation}
\begin{aligned}
&C, F & \leftarrow & \arg\min_{C, F} \mathcal{L}_{Overall}(F,C,D) &\\
&D & \leftarrow & \arg\max_{D} \mathcal{L}_{Overall}(F,C,D) &\\
&\beta^{S} & \leftarrow & \max(0, \beta^{S} + \alpha\mathcal{L}_{ic}^{S}) &\\
&\beta^{T} & \leftarrow & \max(0, \beta^{T} + \alpha\mathcal{L}_{ic}^{T}),&
\label{eq:training}
\end{aligned}
\end{equation}
where $\alpha$ denotes the step length for updating $\beta^S$/$\beta^T$.

\subsection{Significance-aware Information Bottleneck}
\label{SIB}
The starting point for our significance-aware information bottleneck (SIB) is the observation that segmentation of those infrequent classes is prone to be hurt by the standard IB. We analyze the reason from two folds. On the one hand, for the infrequent classes, the supervision is insufficient to support the network to learn a good representation under the constraint from the bottleneck. On the other hand, from the view of information entropy, the actual encoding of an infrequent sample would span more channels in a feature vector. As the $KL$-divergence is calculated by summing up the channel-wise losses, the features from those infrequent classes are usually suffered from more powerful constraint. The problem is severe in semantic segmentation task because the class occupations in a scene are highly unbalanced and the latent features are usually high-dimensional. The proposed SIB aims to address such limitation by incorporating the significance-aware mechanism.

\begin{figure}
\centering
\includegraphics[width=0.9\linewidth]{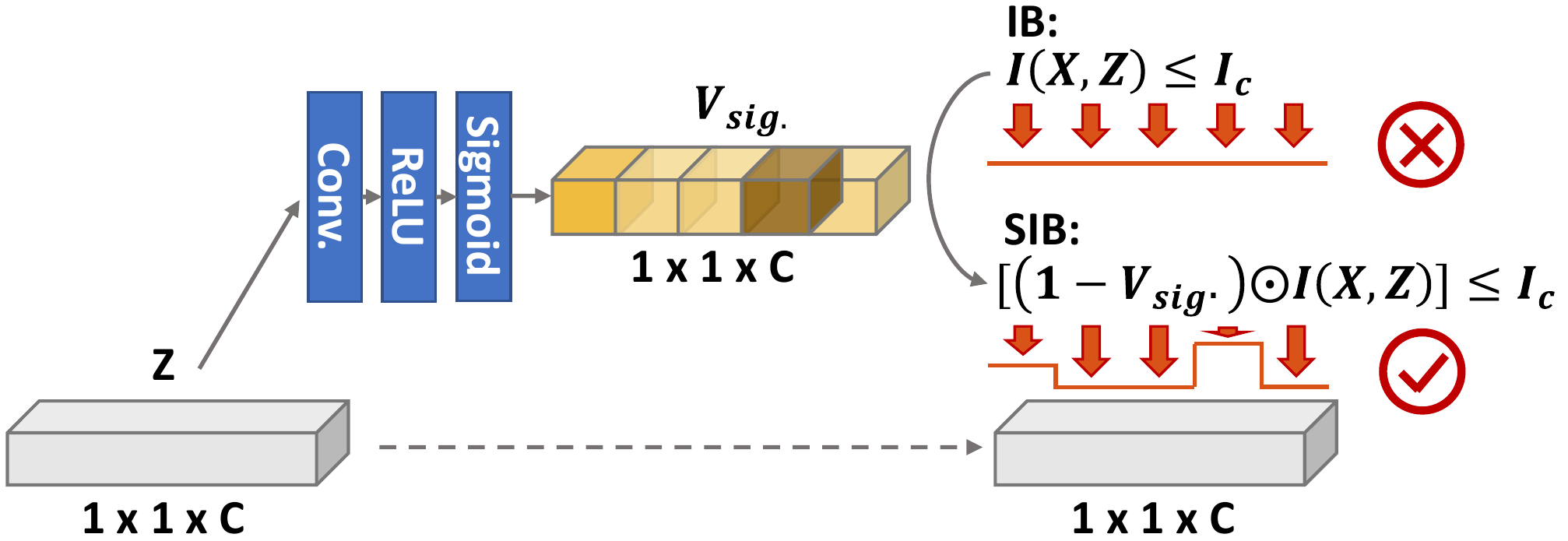}
\caption{Significance-aware information bottleneck (SIB). We use a significance-aware module to detect the channel-wise significance $V_{sig.}$ for each pixel-level feature, with which the original information constraint loss is adaptively weighted. The different sizes of red arrows indicate SIB attaches different compression to each channel according to their significance, while the standard IB compress each channel equally.}
\label{fig:SIB}
\vspace{-0.5cm}
\end{figure}

Fig.~\ref{fig:SIB} details our proposed SIB module. Firstly, we detect the channel-wise significance vector $V_{sig.}$ for the latent feature. Since we adopt a $1 \times 1$ kernel-sized convolutional layer in SIB, here we use a $1\times 1 \times C$ shaped feature vector within the $w\times h \times C$ shaped feature map for illustration. The information constraint is then adaptively weighted by multiplying $1 - V_{sig.}$. Taking the source domain features as an example, the significance-aware IB loss can be obtained as
\begin{equation}
\begin{aligned}
\mathcal{L}_{ic}^{S} = \mathbb{E}_{\mathbf{x} \sim p(\mathbf{x}^S)} [(1 - V_{sig.}) \odot (\mathrm{KL}[F(\mathbf{z}|\mathbf{x}) || r(\mathbf{z})] - I_c)],
\label{eq:significance-aware loss}
\end{aligned}
\end{equation}
where $\odot$ denotes the channel-wise product. The intuition is that the more significant channels should get less constraint.\footnote[1]{It is noteworthy that we do not back-propagate the information constraint loss to the significance-aware layer. Hence the $V_{sig.}$ is only trained to minimize the task loss and does not subject to the IB.} Such SIB can adaptively decrease the constraint effect on important channels, thus preventing the critical information from being eliminated. Experimental results show the proposed SIB brings a significant improvement over the standard IB in segmentation task, especially for those infrequent classes.

\subsection{Network Architecture}
\label{Network}
Our network architecture is illustrated in Fig.~\ref{fig:main}. It is composed of a generator $G$ and a discriminator $D$. $G$ can be any FCN-based segmentation network~\cite{simonyan2014vgg, long2015fcn, chen2018deeplab}, which is further divided into a feature extractor $F$ and a classifier $C$. We attach the SIB on the output from last convolutional layer of $F$. $D$ is a CNN-based binary classifier with a fully-convolutional output~\cite{goodfellow2014gan}, which attempts to distinguish whether a latent feature is from source or target domain.


Given a source domain image and the annotation $(x_i^S, y_i^S)$, $F$ is used to extract a latent representation $z_i^S$ and SIB is applied to $z_i^S$ to conduct the significance-aware feature purification. Specifically, we firstly forward $z_i^S$ to the significance-aware module to yield a channel-wise significance vector $V_{sig.}$ for each pixel-level features. Then $V_{sig.}$ together with $z_i^S$ are fed into IB to calculate a significance-weighted $KL$-divergence between $p(z_i^S)$ and $\mathcal{N}(0; I)$, which is named ``information constraint loss''. Finally, we multiply $z_i^S$ with $V_{sig.}$ to produce $z_{sig}^S$, which denotes the final representation of $x_i^S$. On the one hand, $z_{sig}^S$ is forwarded to $C$ to yield a segmentation loss under the supervision of the ground-truth label $y_i^S$. On the other hand, $z_{sig}^S$ is input to $D$ to generate an adversarial loss. 

Given a target domain image $x_i^T$, we also forward it to $F$ through SIB and obtain a purified latent representation $z_{sig}^T$. Different from the source data flow, since we have no access to the target annotation, we only use the adversarial loss and information constraint loss to train the network. 

\begin{table*}[t]

\caption{
    Adaptation from GTA5~\cite{richter2016gta5} to Cityscapes~\cite{cordts2016cityscapes}. We present the results in terms of per-class IoU and mean IoU. ``F'', ``P'' and ``S'' represent the DA applied in feature space, pixel space and semantic space, respectively. ``VGG-16'' and ``ResNet'' represent the VGG16-FCN8s and ResNet-101 backbones, respectively. IBAN denotes the baseline network equipped with a standard IB. \emph{Gain} indicates the mIoU improvement over using the source only.
    }
\vspace{-0.5cm}
  \begin{center}
  \scriptsize
  \setlength{\tabcolsep}{2.7pt}
  \begin{tabular}{l|c|c|ccccccccccccccccccc>{\columncolor{mygray}}c>{\columncolor{mygray}}c}
    \toprule
    \multicolumn{24}{c}{\textbf{GTA5 $\rightarrow$ Cityscapes}} \\
    \midrule
     &\rot{Space} &\rot{Arch.} & \rot{road} & \rot{side.} & \rot{buil.} & \rot{wall} & \rot{fence} & \rot{pole} & \rot{light} & \rot{sign} & \rot{vege.} & \rot{terr.} & \rot{sky} & \rot{pers.} & \rot{rider} & \rot{car} & \rot{truck} & \rot{bus} & \rot{train} & \rot{motor} & \rot{bike} & \rot{\textbf{mIoU}} & \rot{\textbf{gain}}\\ 
     \midrule
     \midrule
     Source only & - & \multirow{10}{*}{\textbf{\rot{VGG-16}}} & 26.0 & 14.9 & 65.1 & 5.5 & 12.9 &  8.9 &  6.0 &  2.5 & 70.0 &  2.9 & 47.0 & 24.5 &  0.0 & 40.0 & 12.1 &  1.5 &  0.0 &  0.0 &  0.0 & 17.9 & - \\
    
     CyCADA (pixel only)~\cite{hoffman2017cycada} & P &  &  83.5 & 38.3 & 76.4 & 20.6 & 16.5 & 22.2 & 26.2 & 21.9 & 80.4 & 28.7 & 65.7 & 49.4 & 4.2 & 74.6 & 16.0 & 26.6 &2.0 &8.0 & 0.0 & 34.8 & 16.9\\

     AdaptSeg (seg. only)~\cite{tsai2018OutputSpace} & S &  & 87.3 & 29.8 & 78.6 & 21.1 &  18.2 & 22.5 & 21.5 & 11.0 & 79.7 & 29.6 & 71.3 & 46.8 & 6.5 & 80.1 & 23.0 & 26.9 & 0.0 & 10.6 & 0.3 & 35.0 & 17.1\\
    
     \cmidrule{1-2}\cmidrule{4-24}
    Source only & - &  & 26.0 & 14.9 & 65.1 & 5.5 & 12.9 &  8.9 &  6.0 &  2.5 & 70.0 &  2.9 & 47.0 & 24.5 &  0.0 & 40.0 & 12.1 &  1.5 &  0.0 &  0.0 &  0.0 & 17.9 & -\\
    
    FCNs in the wild (feat. only)~\cite{hoffman2016fcns} & F &  & 70.4 & \bf 32.4 & 62.1 & 14.9 &  5.4 & 10.9 & 14.2 &  2.7 & 79.2 & 21.3 & 64.6 & \bf 44.1 &  4.2 & 70.4 &  8.0 &  7.3 &  0.0 &  3.5 &  0.0 & 27.1 & 9.2  \\
    
	CyCADA (feat. only)~\cite{hoffman2017cycada} & F &  & \bf 85.6 & 30.7 & 74.7 & 14.4 & 13.0 & 17.6 & 13.7 & 5.8 & 74.6 & 15.8 & 69.9 & 38.2 & 3.5 & 72.3 & 16.0 & 5.0 & 0.1 & 3.6 & 0.0 & 29.2 & 11.3\\ 
    
    Baseline (feat. only)~\cite{tsai2018OutputSpace} & F &  & 81.8 & 23.5 & 75.2 & 17.6 & 12.8 & 20.3 & 16.9 & \bf 10.8 & 76.4 & 22.6 & 71.3 &  43.8 & 6.5 & 72.1 & 20.0 & \bf 19.5 & 1.2 & 9.6 & 0.3 & 31.7 & 13.8\\

    IBAN (Ours) & F &  & 84.0 & 11.1 & \bf 80.2 & 16.4 & 14.5 & 21.1 & 19.0 & 7.9 & 80.6 & 27.5 & 76.0 & 43.8 & 4.9 & \bf 78.5 & 16.9 & 17.3 & 1.7 & 8.6 & 0.0 & 32.1 & 14.2\\
    
    SIBAN (Ours) & F &  & 83.4 & 13.0 &  77.8 & \bf 20.4 & \bf 17.5 & \bf 24.6 & \bf 22.8 & 9.6 & \bf 81.3 & \bf 29.6 & \bf 77.3 & 42.7 & \bf 10.9 & 76.0 & \bf 22.8 & 17.9 & \bf 5.7 & \bf 14.2 & \bf 2.0 & \bf 34.2 & \bf 16.3\\
		\midrule
        \midrule
        
    Source only & - & \multirow{6}{*}{\textbf{\rot{ResNet}}} & 75.8 & 16.8 & 77.2 & 12.5 & 21.0 & 25.5 & 30.1 & 20.1 & 81.3 & 24.6 & 70.3 & 53.8 & 26.4 & 49.9 & 17.2 & 25.9 & 6.5 & 25.3 & 36.0 & 36.6 & -\\
    
    AdaptSeg (seg. only)~\cite{tsai2018OutputSpace} & S &  & 86.5 & 25.9 & 79.8 & 22.1 & 20.0 & 23.6 & 33.1 &  21.8 & 81.8 & 25.9 & 75.9 & 57.3 & 26.2 & 76.3 & 29.8 & 32.1 & 7.2 & 29.5 & 32.5 & 41.4 & 4.8\\
    
    \cmidrule{1-2}\cmidrule{4-24}
	Source only & F &  & 75.8 & 16.8 & 77.2 & 12.5 & 21.0 & 25.5 & 30.1 & 20.1 & 81.3 & 24.6 & 70.3 & 53.8 & \bf 26.4 & 49.9 & 17.2 & 25.9 & 6.5 & \bf 25.3 & \bf 36.0 & 36.6 & -\\
    
	Baseline (feat. only)~\cite{tsai2018OutputSpace} & F &   & 83.7 & 27.6 & 75.5 & 20.3 & 19.9 & 27.4 & 28.3 & \bf 27.4 & 79.0 & 28.4 & 70.1 & 55.1 & 20.2 & 72.9 & 22.5 & \bf 35.7 & \bf 8.3 & 20.6 & 23.0 & 39.3 & 2.7\\
    
    IBAN (Ours) & F &   & 88.2 & 33.7 & \bf 80.1 & 23.4 & 21.8 & 27.7 & 27.9 & 16.3 & \bf 83.2 & 38.3 & 76.2 & 57.5 & 20.3 & 81.1 & 25.9 & 33.4 & 1.9 & 22.4 & 20.7 & 40.7 & 4.1\\ 
    
	SIBAN (Ours)  & F &   & \bf 88.5 & \bf 35.4 & 79.5 & \bf 26.3 & \bf 24.3 & \bf 28.5 & \bf 32.5 & 18.3 & 81.2 & \bf 40.0 & \bf 76.5 & \bf 58.1 & 25.8 & \bf 82.6 & \bf 30.3 & 34.4 & 3.4 & 21.6 & 21.5 & \bf 42.6 & \bf 6.0\\

    \bottomrule
  \end{tabular}
  \end{center}
  \label{table:gta5-cityscapes}
\end{table*}

\begin{table*}[t]
\caption{
    Adaptation from Synthia~\cite{ros2016synthia} to Cityscapes~\cite{cordts2016cityscapes}. The table setting is the same as Table~\ref{table:gta5-cityscapes}.
    }
  \begin{center}
  \scriptsize
  \setlength{\tabcolsep}{5.55pt}
  \begin{tabular}{l|c|c|ccccccccccccc>{\columncolor{mygray}}c>{\columncolor{mygray}}c}
    \toprule
    \multicolumn{18}{c}{\textbf{SYNTHIA $\rightarrow$ Cityscapes}} \\
    \midrule
     &\rot{Space} &\rot{Arch.} & \rot{road} & \rot{side.} & \rot{buil.} &   \rot{light} & \rot{sign} & \rot{vege.} & \rot{sky} & \rot{pers.} & \rot{rider} & \rot{car}  & \rot{bus} & \rot{motor} & \rot{bike} & \rot{\textbf{mIoU}} & \rot{\textbf{gain}}\\ 
     \midrule
     \midrule
     Source only & - & \multirow{9}{*}{\textbf{\rot{VGG-16}}} & 6.4 & 17.7 & 29.7 & 0.0 & 7.2 & 30.3 &  66.8 &  51.1 & 1.5 &  47.3 & 3.9 & 0.1 & 0.0 & 20.2 & -\\

     AdaptSeg (seg. only)~\cite{tsai2018OutputSpace} & S &  & 78.9 & 29.2 & 75.5 & 0.1 & 4.8 & 72.6 & 76.7 & 43.4 & 8.8 & 71.1 & 16.0 & 3.6 & 8.4 & 37.6 & 17.4\\
    
     \cmidrule{1-2}\cmidrule{4-18}
     Source only & - &  & 6.4 & 17.7 & 29.7 & 0.0 & 7.2 & 30.3 &  66.8 & \bf 51.1 & 1.5 &  47.3 & 3.9 & 0.1 & 0.0 & 20.2 & -\\
    
     FCNs in the wild (feat. only)~\cite{hoffman2016fcns} & F &  & 11.5 & 18.3 & 33.3 & 0.0 & \bf 11.2 & 43.6 & 70.5 & 45.5 & 1.3 & 45.1 & 4.6 & 0.1 & 0.5 & 22.0 & 1.8 \\
    
	 Cross-city (feat. only)~\cite{hoffman2017cycada} & F &  & 56.5 & 24.0 & 78.9 & 1.1 & 5.9 & \bf 77.8 & 77.3 & 35.8 & 5.4 & 61.7 & 5.2 & 0.9 & 8.4 & 33.8 & 13.6 \\ 
    
     Baseline (feat. only)~\cite{tsai2018OutputSpace} & F &  & 63.1 & 17.9 & 76.3 & \bf 4.7 & 8.4 & 68.3 & 79.9 & 38.7 & \bf 8.5 & 64.7 & 9.7 & 0.6 & 6.0 & 34.4 & 14.2 \\
     
     IBAN (Ours) & F &  & 70.0 & 19.1 & 78.7 & 1.4 & 4.5 & 73.1 & 77.0 & 42.2 & 2.6 & 72.5 & 14.0 & 0.8 & 3.9 & 35.4 & 15.2 \\
    
     SIBAN (Ours) & F &  & \bf 70.1 & \bf 25.7 & \bf 80.9 & 3.8 & 7.2 & 72.3 & \bf 80.5 & 43.3 & 5.0 & \bf 73.3 & \bf 16.0 & \bf 1.7 & \bf 3.6 & \bf 37.2 & \bf 17.0 \\
		
		\midrule
        \midrule
        
     Source only & - & \multirow{6}{*}{\textbf{\rot{ResNet}}} & 55.6 & 23.8 & 74.6 & 6.1 & 12.1 & 74.8 & 79.0 & 55.3 & 19.1 & 39.6 & 23.3 & 13.7 & 25.0 & 38.6 & -\\
    
     Baseline (seg. only)~\cite{tsai2018OutputSpace} & S &  & 79.2 & 37.2 & 78.8 & 9.9 & 10.5 & 78.2 & 80.5 & 53.5 & 19.6 & 67.0 & 29.5 & 21.6 & 31.3 & 45.9 & 7.3\\
    
     \cmidrule{1-2}\cmidrule{4-18}
	 Source only & F &  & 55.6 & 23.8 & 74.6 & 6.1 & 12.1 & 74.8 & 79.0 & 55.3 & \bf 19.1 & 39.6 & 23.3 & 13.7 & 25.0 & 38.6 & -\\
    
	 Baseline (feat. only)~\cite{tsai2018OutputSpace} & F &   & 62.4 & 21.9 & 76.3 & 11.7 & 11.4 & 75.3 & 80.9 & 53.7 & 18.5 & 59.7 & 13.7 & \bf 20.6 & 24.0 & 40.8 & 2.2\\
    
     IBAN (Ours)  & F &   & 78.2 & 19.7 &  \bf 80.5 & 9.4 & 8.9 & 77.4 & 82.0 & 56.3 & 9.6 & 76.3 & 22.8 & 17.5 & 23.3 & 43.2 & 4.6 \\ 
    
	 SIBAN (Ours) & F &   & \bf 82.5 & \bf 24.0 & 79.4 & \bf 16.5 & \bf 12.7 & \bf 79.2 & \bf 82.8 & \bf 58.3 & 18.0 & \bf 79.3 & \bf 25.3 & 17.6 & \bf 25.9 & \bf 46.3 & \bf 7.7\\ 

    \bottomrule
  \end{tabular}
  \end{center}
  \label{table:synthia-cityscapes}
\end{table*}

\subsection{Theoretical Insight}
\label{Analysis}
In this section, we show the relationship between our method and the theory of domain adaptation proposed by Ben-David \emph{et al}~\cite{ben2010theory}. 

\textbf{Theorem 1.} Let $\mathcal{H}$ be the hypothesis class, $S$ and $T$ denote two different domains, we have the theory bounds the expected error on the target samples $\epsilon^T(h)$ by three terms as follows:
\vspace{-0.1cm}
\begin{equation}
\begin{aligned}
\forall h \in \mathcal{H}, \epsilon^T(h) \leq \epsilon^S(h) + \frac{1}{2} d_{\mathcal{H}\Delta\mathcal{H}}(S, T) + \lambda,
\label{eq:theorem}
\end{aligned}
\end{equation}
where $\epsilon^S(h)$ is the expected error on the source samples
which can be minimized easily in a fully-supervised manner, $d_{\mathcal{H}\Delta\mathcal{H}}(S, T)$ denotes a discrepancy distance between source and target distributions \emph{w.r.t.} a hypothesis set $\mathcal{H}$. $\lambda$ is the shared expected loss and is expected to be negligibly small. 

Without loss of generality, we represent the source feature as a concatenation of task-dependent part $z$ and task-independent part $\hat{z}$. Since the $z$ and $\hat{z}$ are expected as mutual independent, we can rewrite the source distribution $S$ as $S = S_z \times S_{\hat{z}}$, where $S_z$ and $S_{\hat{z}}$ denote the marginal distribution of $z$ and $\hat{z}$. We use the parallel notation $T = T_z \times T_{\hat{z}}$ for the target domain.

Using the results proved in~\cite{talagrand1995concentration} for concentration of measures in product spaces, we can upper bound the $\mathcal{H}\Delta\mathcal{H}$-distance between $S$ and $T$ as follows:

\begin{equation}
\begin{aligned}
d_{\mathcal{H}\Delta\mathcal{H}}(S, T)& =  d_{\mathcal{H}\Delta\mathcal{H}}(S_z \times S_{\hat{z}}, T_z \times T_{\hat{z}}) \\
& \leq d_{\mathcal{H}\Delta\mathcal{H}}(S_z, T_z) + \int d_{\mathcal{H}\Delta\mathcal{H}}(S_{\hat{z}}|S_z, T_{\hat{z}})\mathbf{d}S_z
\label{eq:proof}
\end{aligned}
\end{equation}

Recall that our method (see Eq.~\ref{eq:constraint}) enforces the distribution of task-independent factors to approximate a standard Gaussian: $(S_{\hat{z}}|S_z) = S_{\hat{z}} \rightarrow \mathcal{N}(0; I)$ and $T_{\hat{z}} \rightarrow \mathcal{N}(0; I)$, hence enforcing the last integral term of Eq.~\ref{eq:proof}: $d_{\mathcal{H}\Delta\mathcal{H}}(S_{\hat{z}}|S_z, T_{\hat{z}}) \rightarrow d_{\mathcal{H}\Delta\mathcal{H}}(\mathcal{N}, \mathcal{N}) = 0$. Consequently, our method attempts to optimize
the upper bound for $d_{\mathcal{H}\Delta\mathcal{H}}(S, T)$, thus offering a tighter upper bound for $\epsilon^T(h)$. The proof shows that our method is mathematically consistent with the theory of Ben-David \emph{et al}~\cite{ben2010theory}.


\section{Experiments}
\subsection{Datasets} We evaluate our algorithm together with several state-of-the-art algorithms on two adaptation tasks, \emph{e.g.}, SYNTHIA~\cite{ros2016synthia} $\rightarrow$ Cityscapes~\cite{cordts2016cityscapes} and GTA5~\cite{richter2016gta5} $\rightarrow$ Cityscapes. Cityscapes is a real-world dataset with $5,000$ street scenes which are divided into a training set with $2,975$ images, a validation set with $500$ images and a testing set with $1,525$ images. We use Cityscapes as the target domain. GTA5 contains $24,966$ high-resolution images, automatically annotated into $19$ classes. The dataset is rendered from a modern computer game, Grand Theft Auto V, whose labels are fully compatible with Cityscapes. SYNTHIA contains $9,400$ synthetic images compatible with the Cityscapes annotated classes. We use SYNTHIA or GTA5 as the source domain in the evaluation. 

\begin{figure*}
\begin{minipage}[b]{.33\textwidth}
\includegraphics[height=3.2cm]{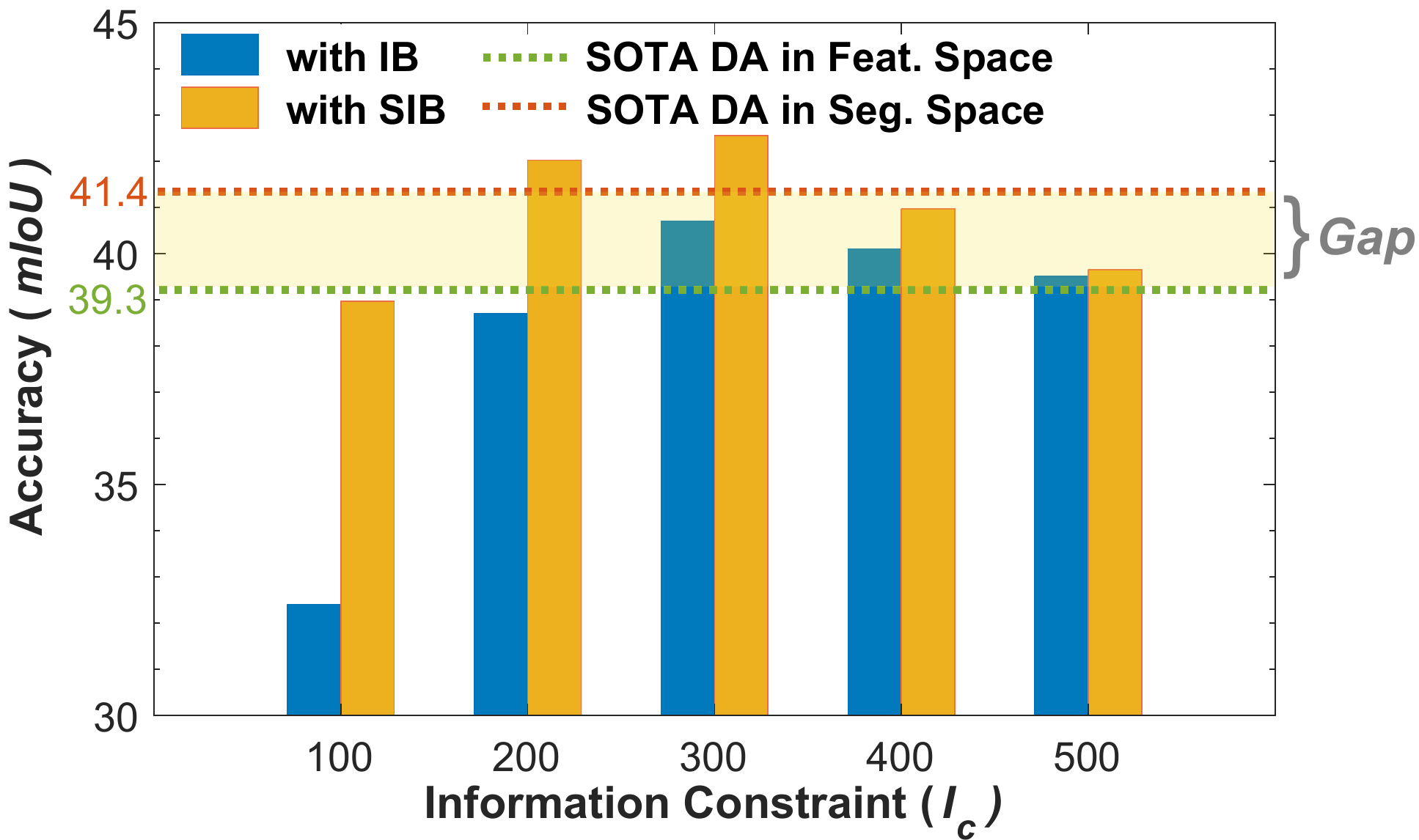}
\subcaption{}
\label{fig:over_the_gap}
\end{minipage}
\begin{minipage}[b]{.33\linewidth}
\centering
\includegraphics[height=3.2cm]{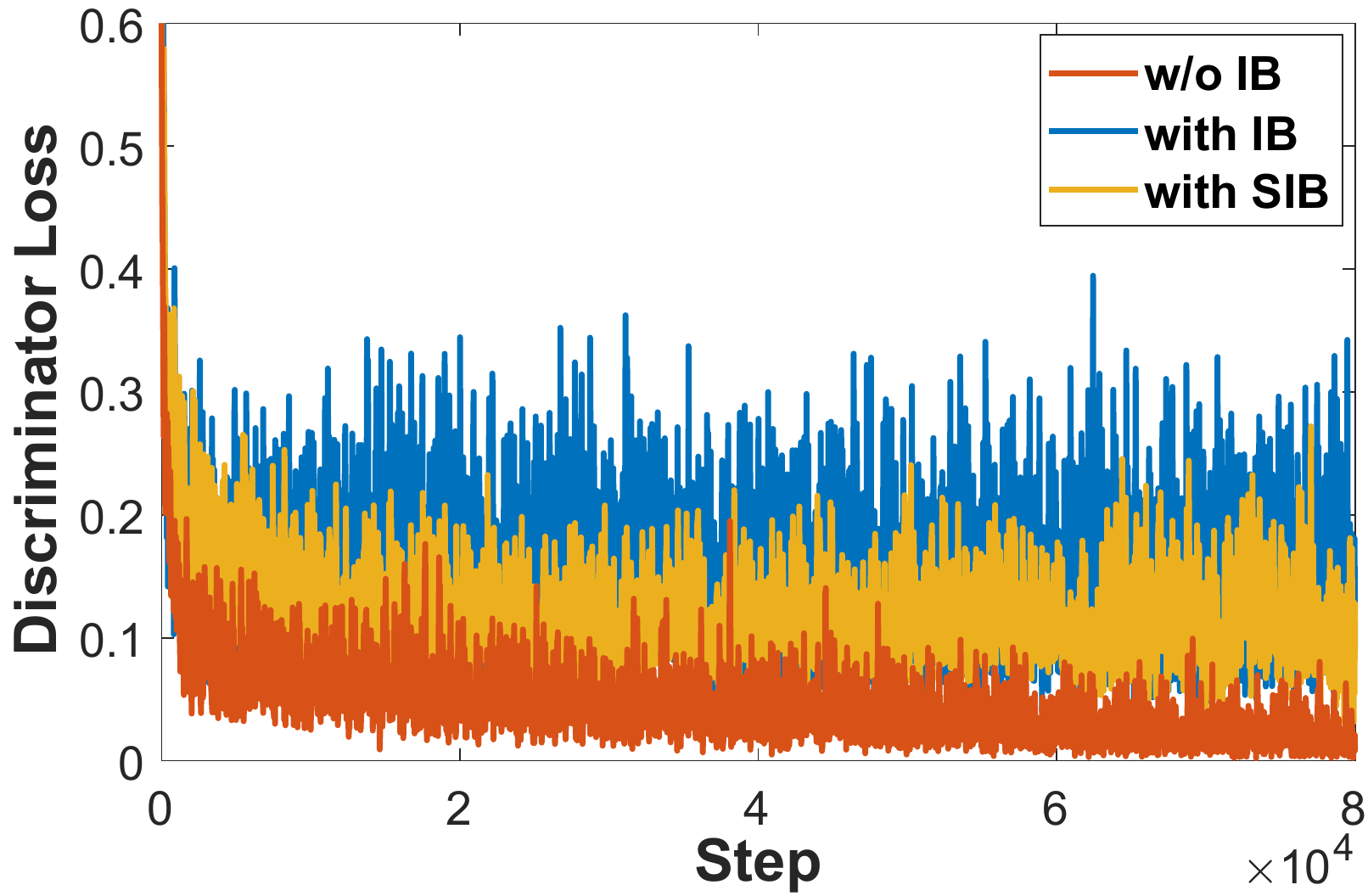}
\subcaption{}
\label{fig:D_loss}
\end{minipage}
\begin{minipage}[b]{.33\linewidth}
\centering
\includegraphics[height=3.2cm]{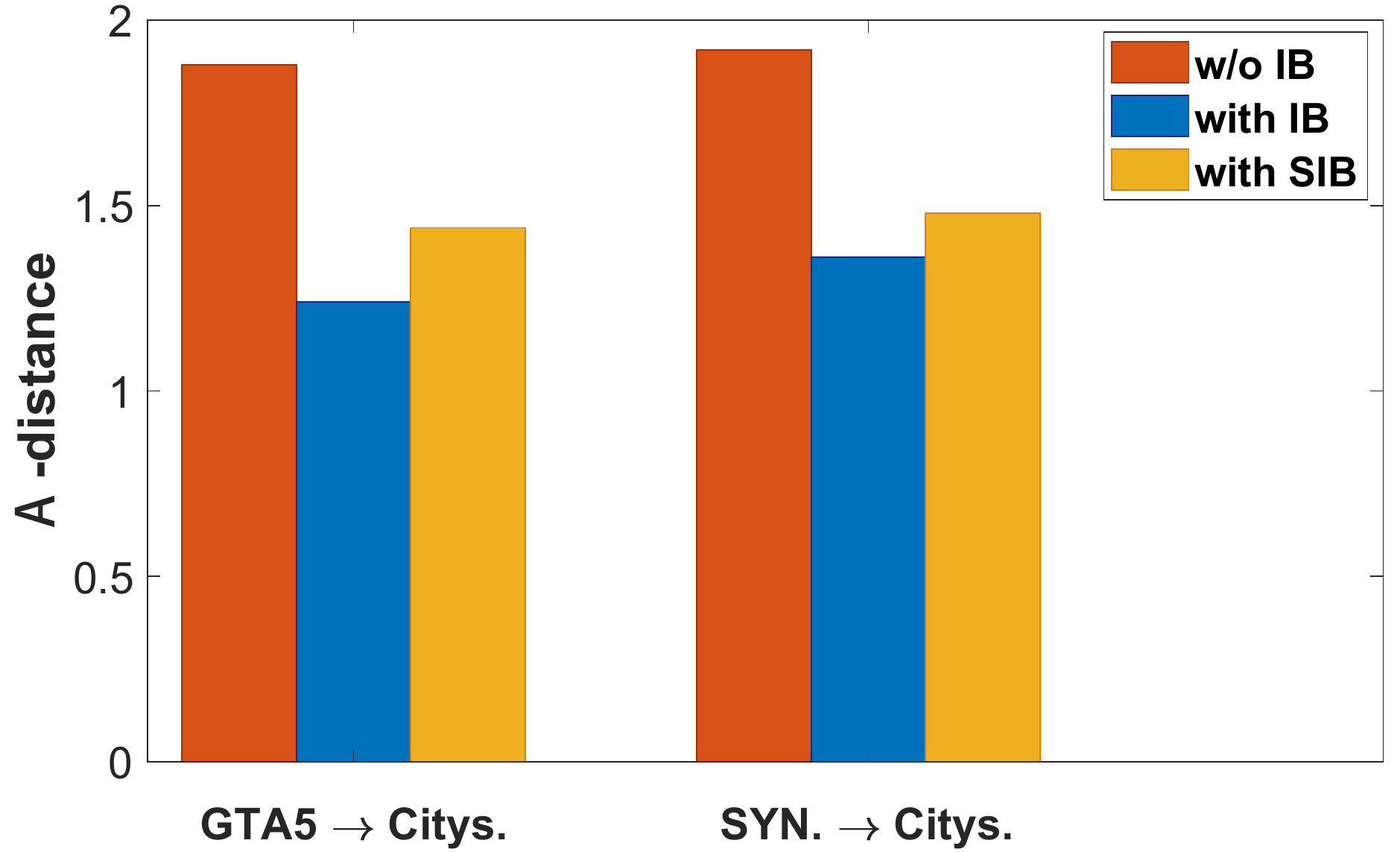}
\subcaption{}
\label{fig:A_distance}
\end{minipage}
\caption{(a). Adapted segmentation performance in terms of mIoU. (b). The training loss of $D$, where a complete balanced adversarial process is achieved when the loss converges to around $0.5$. (c). $\mathcal{A}$-distance between source and target domain.}
\label{fig:three}
\end{figure*}

\subsection{Implementation Details}
We use PyTorch for our implementation. We utilize 1) DeepLab-v2~\cite{chen2018deeplab} framework with ResNet-101~\cite{he2016resnet} and 2) VGG-16-based FCN8s~\cite{long2015fcn}, as the two respective backbones for $G$. We use the feature-space adversarial DA method proposed in~\cite{tsai2018OutputSpace} as the baseline network. For significance-aware layer in SIB, we employ a convolution layer with kernel $1 \times 1$ and channel number $2,048$, followed by a ReLU and a Sigmoid to produce the channel-wise significance vector. We use the IB proposed in~\cite{alexander2017variational} as our bottleneck module. For network $D$, we adopt a similar structure with~\cite{radford2015unsupervised}, which consists of 5 convolution layers with channel numbers $\{64, 128, 256, 512, 1\}$,  the kernel $4 \times 4$ , and stride of $2$. Each convolution layer is followed by a Leaky-ReLU~\cite{maas2013ReLU} parameterized by $0.2$ except the last layer. 
During training, we use SGD~\cite{bottou2010SGD} as the optimizer for $G$ with a momentum of $0.9$, while using Adam~\cite{kingma2014adam} to optimize $D$ with $\beta_1 = 0.9$, $\beta_2 = 0.99$. We set both optimizers a weight decay of $5e$-$4$. The initial learning rates for SGD and Adam are set to $2.5e$-$4$ and $1e$-$4$, respectively. Both learning rate are decayed by a poly policy, where the initial learning rate is multiplied by $(1 - \frac{iter}{max\_iter})^{power}$ with $power = 0.9$. We train the network for a total of $100k$ iterations. We use a crop of $512 \times 1,024$ during training, and for evaluation we up-sample the prediction map by a factor of 2 and then evaluate mIoU. In our best model, we set hyper-parameters $\beta^S_{init} = \beta^T_{init} = 1e$-$5$, $\alpha = 1e$-$8$, $\lambda = 1e$-$3$ and $I_c = 300$, respectively.

\begin{table}[t]
\caption{Ablation study on ResNet-101.}
\label{tab:ablation}
\centering
\begin{tabular}{cc|cc|>{\columncolor{mygray}}c}
\hline
\multicolumn{5}{c}{\textbf{GTA5 $\rightarrow$ Cityscapes}}\\
\hline
\multicolumn{2}{c|}{\textbf{Module}} & \multicolumn{2}{c|}{\textbf{Extra $D$}} & \textbf{mIoU}\\
$SA-layer$ & $Ada. \beta$  & $Sig.~\cite{kang2018deep}$ & $Seg.~\cite{tsai2018OutputSpace}$ & \\ 
\hline
        &			&			&			&   $40.7$ \\
$\surd$	&			&			&			&	$42.2$ \\
$\surd$	& $\surd$	&			&			&	$42.6$ \\
$\surd$	& $\surd$	& $\surd$	&			&	$43.2$ \\
$\surd$	& $\surd$	& $\surd$	& $\surd$	&	$\mathbf{45.5}$ \\

\hline
\end{tabular}
\end{table}

\subsection{Comparative Studies}
\begin{figure*}[ht]
\centering
\includegraphics[width=0.95\linewidth]{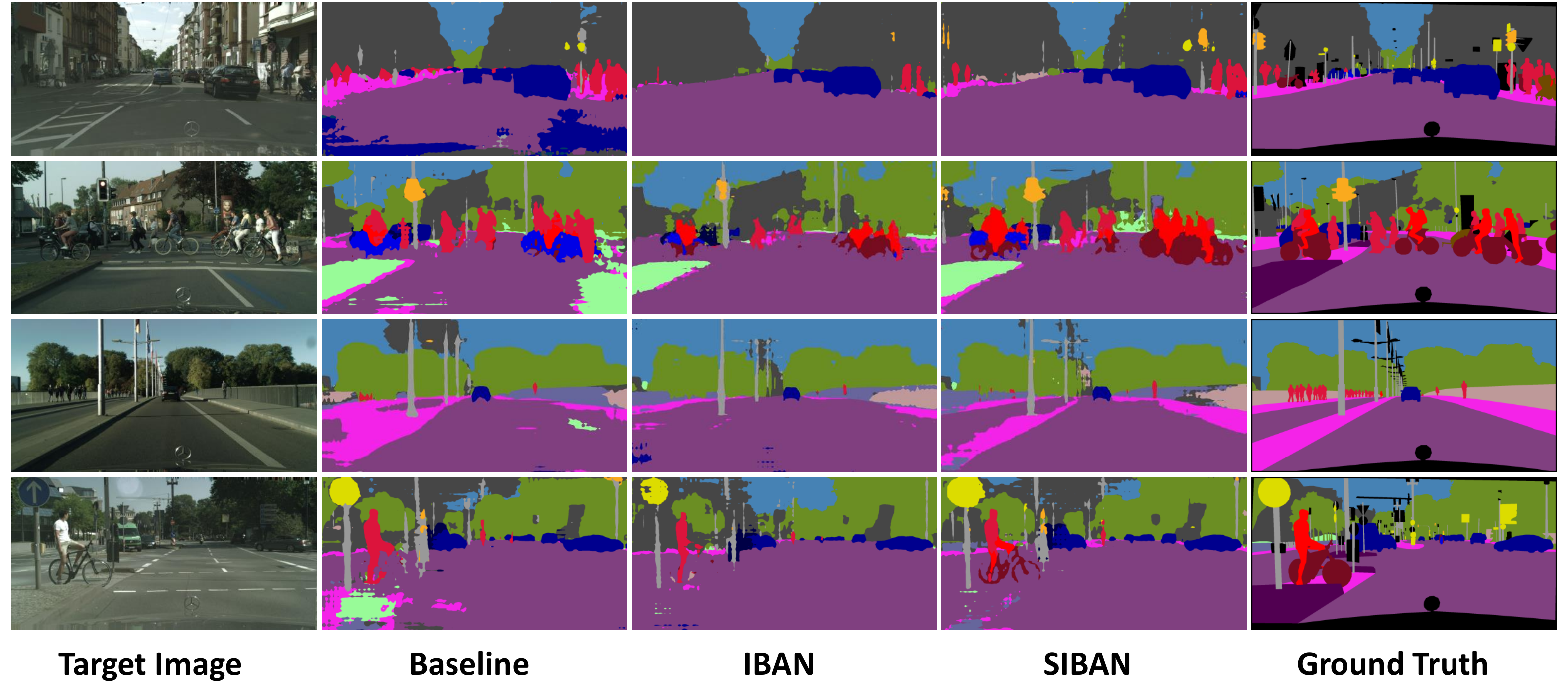}
\caption{Qualitative results of UDA segmentation for GTA5 $\rightarrow$ Cityscapes. For each target image, we show the adapted result with baseline model, IBAN and SIBAN respectively, followed by the ground truth label map. We can find that both IBAN and SIBAN outperform baseline method, and SIBAN can protect the uncommon classes from being eliminated.}
\label{fig:result}
\end{figure*}

\textbf{Compared with SOTA.}
We present the adaptation results on tasks GTA5 $\rightarrow$ Cityscapes and SYNTHIA $\rightarrow$ Cityscapes in Table~\ref{table:gta5-cityscapes} and Table~\ref{table:synthia-cityscapes} respectively, with comparisons to the state-of-the-art feature-space DA methods~\cite{hoffman2016fcns,hoffman2017cycada,tsai2018OutputSpace,chen2017cross}. We also present the current state-of-the-art pixel-space and segmentation-space DA in the tables. In Table~\ref{table:gta5-cityscapes}, not surprisingly, SIBAN significantly outperforms the source-only segmentation method by $+16.3\%$ on VGG-16 and $+6.0\%$ on ResNet-101 since the source-only segmentation method does not consider the domain shift. Besides, SIBAN outperforms the state-of-the-art feature-space methods, which improves the mIOU by over $+2.5\%$ compared with FCNs~\cite{hoffman2016fcns}, AdaptSeg~\cite{tsai2018OutputSpace}, and CyCADA~\cite{hoffman2017cycada}. Compared to the DA methods in the segmentation and pixel space~\cite{tsai2018OutputSpace,hoffman2017cycada}, SIBAN can also be on par with them. In some infrequent classes which are prone to suffer from the side effect of information bottleneck, \emph{e.g.,} fence, traffic light, and rider, we can observe that SIBAN can significantly outperform IBAN. The results verify the effectiveness of SIB module to protect the uncommon classes from being eliminated. Similar results can be observed in Table~\ref{table:synthia-cityscapes}. Some qualitative segmentation examples can be viewed in Fig.~\ref{fig:result}.

\textbf{Sensitivity to Constraint.}
We test the DA performance of IBAN / SIBAN in term of mIoU with varying $I_c$ over a range \{100, 200, 300, 400, 500\}, where a smaller $I_c$ indicating a more strict information constraint on the latent features. Fig.~\ref{fig:over_the_gap} presents the test results, in which we can see that SIBAN outperforms IBAN in all constraint cases. For SIBAN, the appropriate choice of $I_c$ is between $200$ and $400$. An $I_C$ with too small value would eliminate too much essential information, while an excessively large $I_c$ would degrade SIBAN to the baseline model since it introduces too much noise. We can also observe that the IBAN is more sensitive to the constraint. When using $I_c = 300$, both IBAN and SIBAN surpass the feature-space baseline significantly and \textbf{SIBAN can even outperform the state-of-the-art segmentation-space DA methods~\cite{tsai2018OutputSpace}.} From the result, we can conclude that our proposed SIBAN has bridged the performance gap between feature-space and segmentation-space DA methods~\cite{tsai2018OutputSpace}.

\textbf{Training Stability.} Here we utilize the loss of $D$ ($Loss_D$) as a proxy for the stability of adversarial training. In a stable adversarial course, $G$ would learn to fool $D$ successfully, and $Loss_D$ should converge to around 0.5. Fig.~\ref{fig:D_loss} reports $Loss_D$ over the course of training. We can see that $Loss_D$ quickly drops when the network is trained without IB, indicating $D$ overpowers $G$ substantially and learns to differentiate between features of the two domains accurately. We also observe that the introduction of IB / SIB into the adversarial network can significantly constrain the performance of $D$, thus stabilizing the adversarial training. Besides, we find the standard IB outperforms SIB, which seems contradictory to our standpoint. We ascribe it to the reason that a standard IB eliminates excessive information from features. Although making the training of $D$ more stable, such comparatively less-informative features would also hurt the semantic segmentation task. On the contrary, our proposed SIB module can achieve both good training stability and outstanding segmentation performance.

\textbf{$\mathcal{A}$-distance.} Based on the theory of Ben-David \emph{et al.}~\cite{ben2010theory}, $\mathcal{A}$-distance is used as a metric for the domain discrepancy, where a smaller $\mathcal{A}$-distance might indicate better DA performance. Generally, the $\mathcal{A}$-distance is computed as $d_{\mathcal{A}} = 2(1-2\epsilon)$, where $\epsilon$ is the generalization error of a classifier trained with the binary classification task of discriminating the source and target. In the adversarial training framework, we can just keep $D$ as such a classifier. The comparative results are shown in Fig.~\ref{fig:A_distance}. From this figure, we can see that the introduction of IB / SIB significantly reduces $\mathcal{A}$-distances compared to the baseline. However, we can also observe that the $\mathcal{A}$-distance of IBAN is slightly smaller then SIBAN. Consistent with our previous analysis on training stability, we conclude that the discrepancy decrease of IB is at the cost of discarding some necessary information. This finding tells us that only reducing the global distribution discrepancy is far from enough for domain adaptation. The superior DA performance, as well as a relatively small $\mathcal{A}$-distance lead by SIBAN, show that our method can make a better trade-off between the feature purification and domain alignment.

\subsection{Ablation Studies}
To assess the importance of various aspects of the model, we run experiments on GTA5 $\rightarrow$ Cityscapes task on ResNet-101 backbone, deactivating one or a few modules at a time while keeping the others activated. Besides, we test the combination performance between SIBAN and other DA methods~\cite{tsai2018OutputSpace,kang2018deep}, in which the author suggests the channel-wise significance~\cite{kang2018deep} or the output~\cite{tsai2018OutputSpace} should also be aligned between domains. We simply implement these two methods by adding two extra discriminators $D$ on significance tensors and segmentation maps, respectively. Table~\ref{tab:ablation} shows the DA results under different settings. We observe that appending SA-layer can significantly improve the standard IB by $1.5\%$. Updating $\beta_S/\beta_T$ adaptively brings extra $0.4\%$ improvement as well. When employing two extra discriminators to the significance tensors and the segmentation maps, the target segmentation accuracy would be further improved by $0.8\%$ and $2.3\%$. The ablation study verifies the effectiveness of our SIB module as well as our ``adaptive $\beta$'' strategy for DA task. Furthermore, SIBAN can be expediently combined with other DA methods to yield even better segmentation results on target images.

\section{Conclusion}
In this paper, we propose a novel significance-aware information bottlenecked adversarial network (SIBAN) for domain adaptive semantic segmentation. By conducting a significance-aware feature purification before the adversarial adaptation, SIBAN eases the following feature alignment and stabilizes the adversarial training course, thus significantly improving the feature-space adaptation performance. On two challenging $similated \rightarrow real$ DA tasks, SIBAN yields leading result compared with other feature-space methods, and can even match the state-of-the-art output-space methods in segmentation accuracy. For the semantic segmentation task, our proposed SIBAN brings the feature-/output-space UDA methods to the same starting line.

{\small
\bibliographystyle{ieee}
\bibliography{egpaper_for_review}
}

\clearpage

\vspace{0.5cm}
\twocolumn[ \huge \textbf{\emph{Appendix}}\vspace{1cm}]

\section*{Hyper-parameter Analysis}
\begin{figure}[t]
    \centering
    \includegraphics[width=0.95\linewidth]{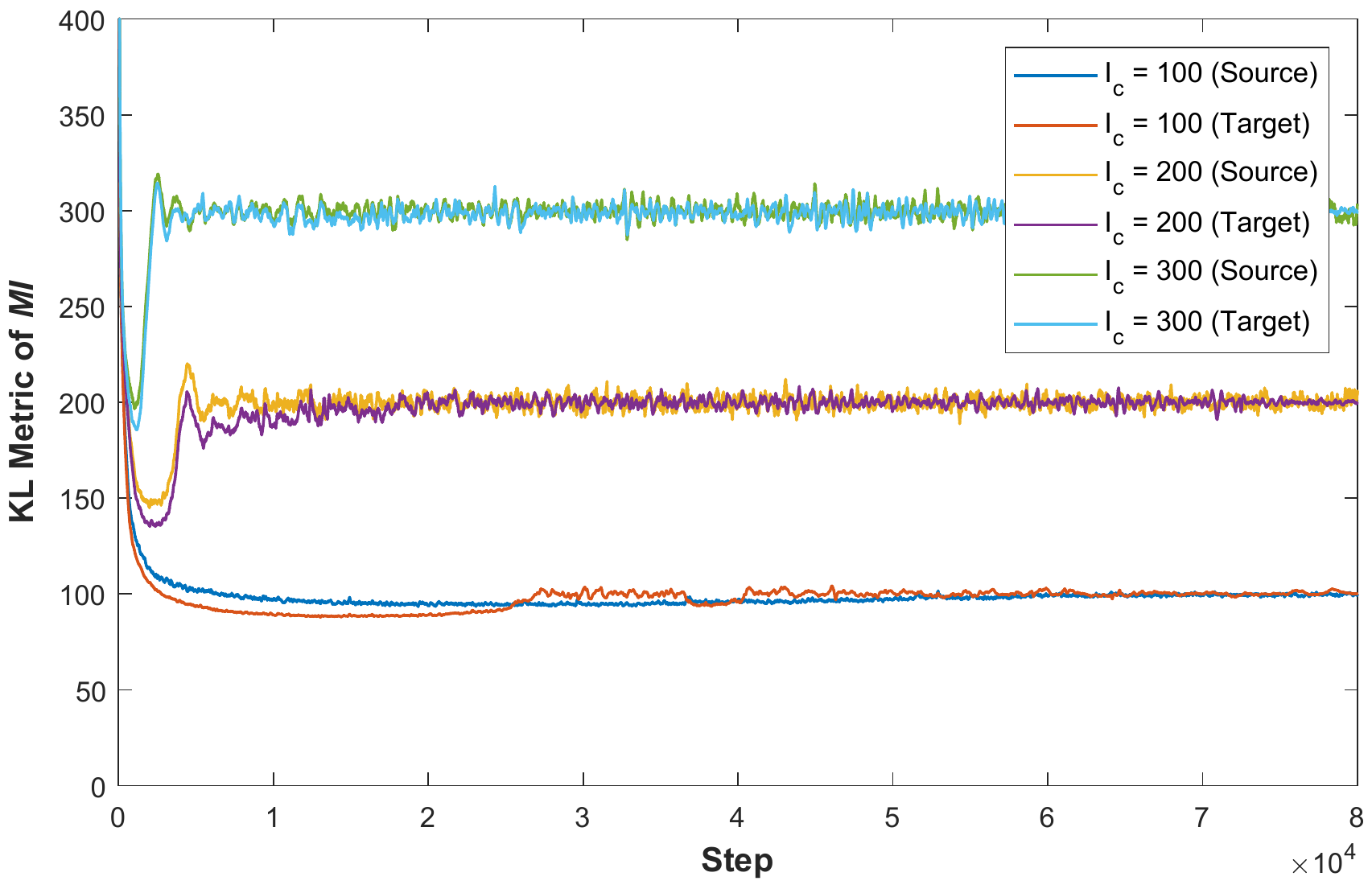}
    \caption{$KL$-divergence curves over the course of training.}
    \label{fig:KL}
    \vspace{0.5cm}
\end{figure}

\begin{figure}[t]
    \centering
    \includegraphics[width=0.94\linewidth]{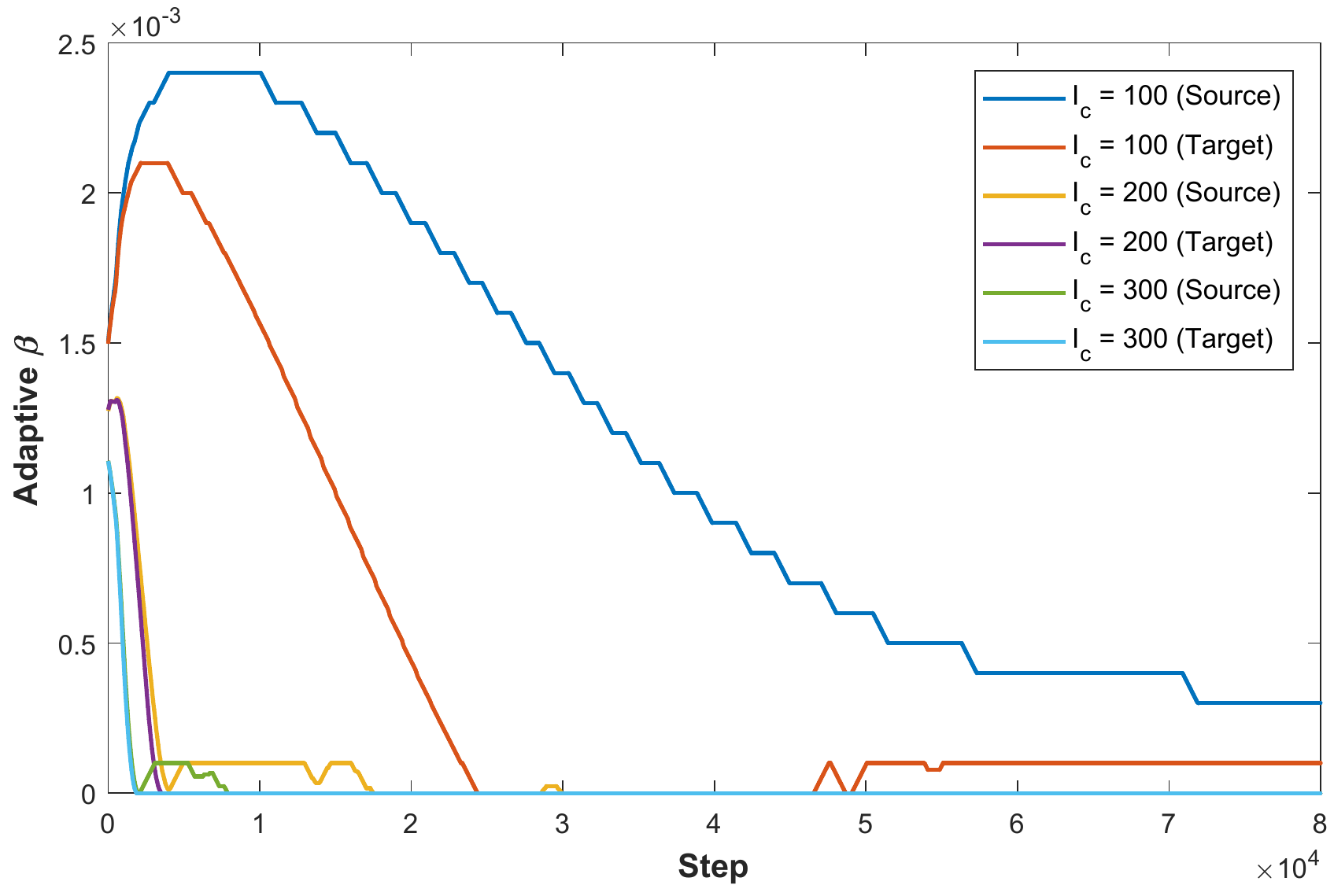}
    \caption{The values of adaptive $\beta_S / \beta_T$ over the course of training.}
    \label{fig:beta}
    \vspace{0.4cm}
\end{figure}

The information constraint $I_c$ is an important hyper-parameter in SIBAN to control the feature purification. Here we train SIBAN with varying $I_c$ over a range \{100, 200, 300\}, and report the corresponding KL-divergence curves for both domains in Fig.~\ref{fig:KL}, as well as the corresponding adaptive $\beta_S / \beta_T$ values over the training course in Fig.~\ref{fig:beta}. From the results, we can see that the information can be easily constrained to a specific $I_c$ when choosing a relatively large $I_c$ between $[200,300]$. When choosing $I_c = 100$, the feature purification process becomes harder, since an information bottleneck with such a small $I_c$ ($100$) is too narrow to maintain the necessary information for segmentation (see Fig.~\ref{fig:KL}). Accordingly, the model has to give more bias to decrease the information constraint loss when choosing a small $I_c$, which explains why the $\beta_S / \beta_T$ are relative large during the training course (see Fig.~\ref{fig:beta}). We also observe that the information from the target domain is easier to be constrained to $I_c$. This is because the target model is trained under an unsupervised mechanism, which is more easily dominated by the information constraint loss.

\section*{Feature Distribution Visualization}
\begin{figure*}[htbp]
\begin{minipage}[b]{.35\textwidth}
\centering
\includegraphics[height=4cm]{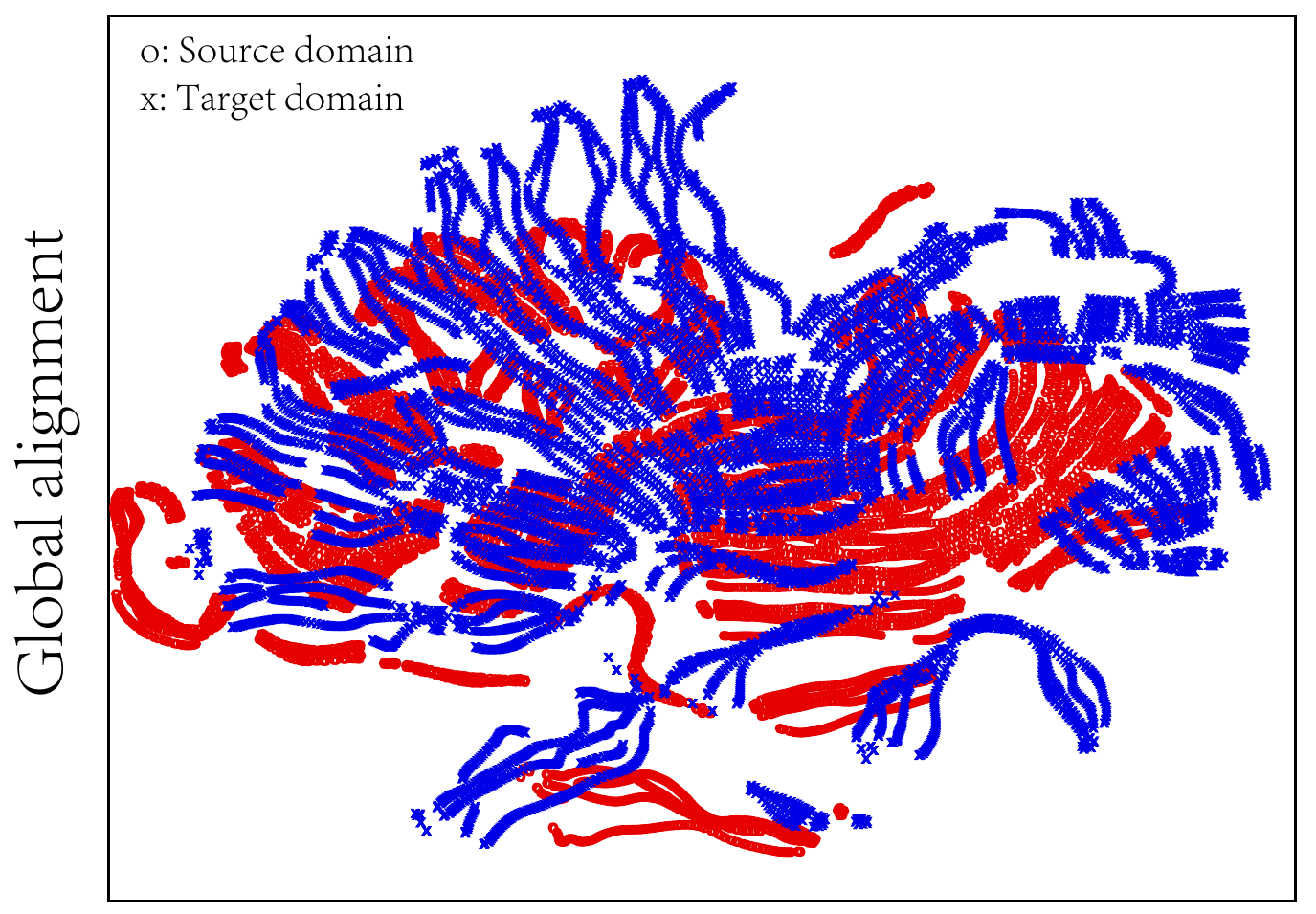}
\subcaption{Non-adapted}
\label{fig:tsne_NoAN_Domain}
\end{minipage}
\begin{minipage}[b]{.32\linewidth}
\centering
\includegraphics[height=4cm]{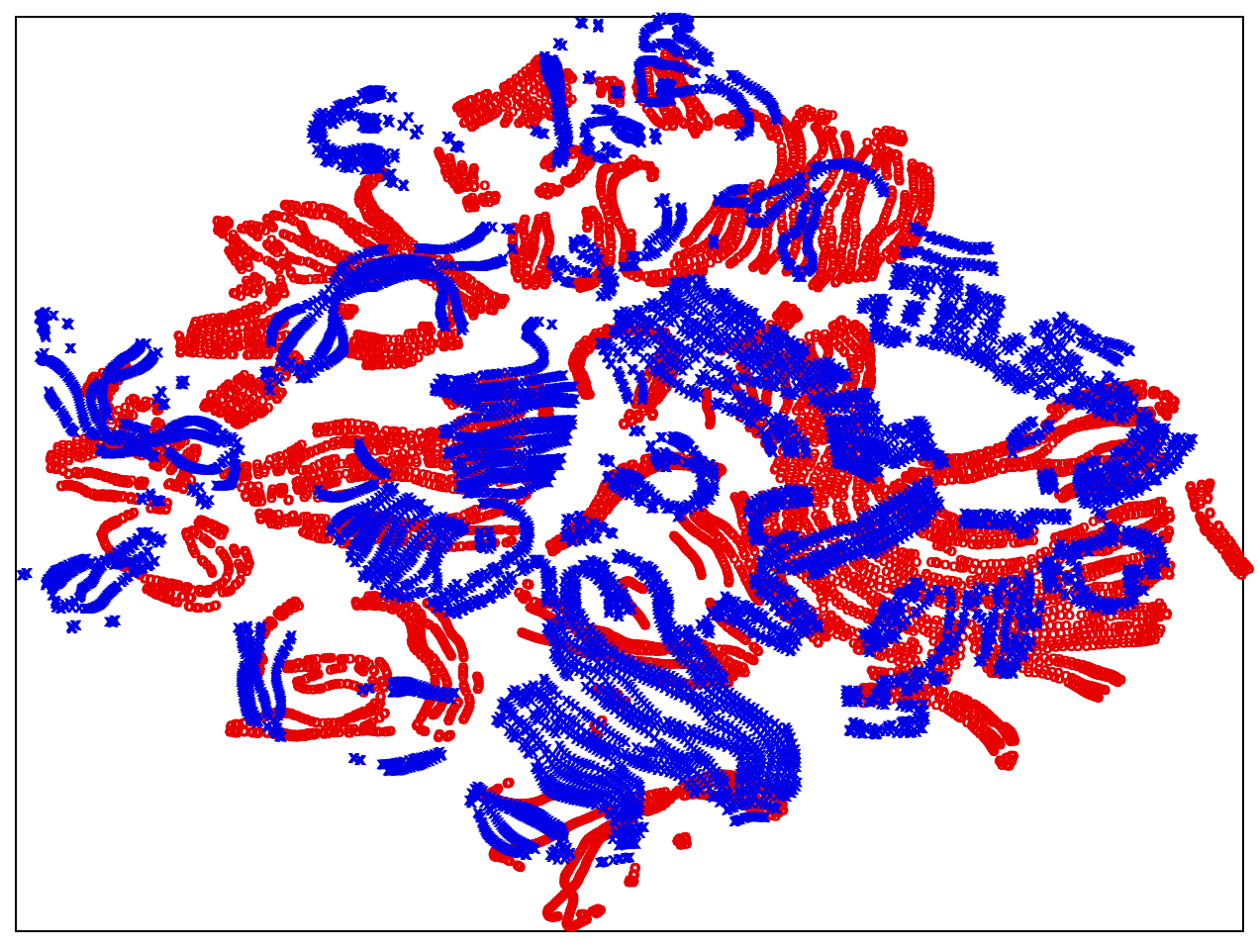}
\subcaption{IBAN}
\label{fig:tsne_IBAN_Domain}
\end{minipage}
\begin{minipage}[b]{.32\linewidth}
\centering
\includegraphics[height=4cm]{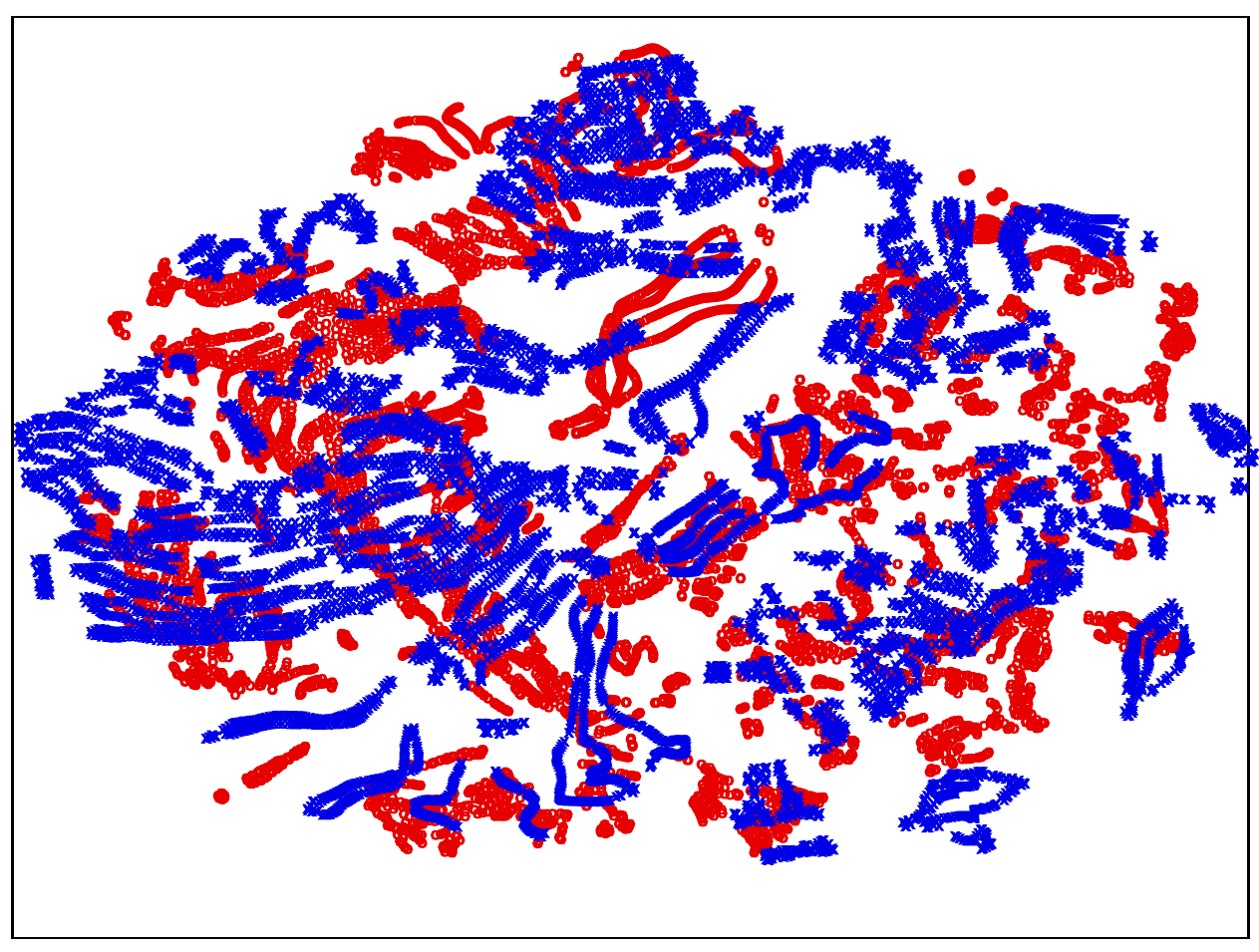}
\subcaption{SIBAN}
\label{fig:tsne_SIBAN_Domain}
\end{minipage}

\begin{minipage}[b]{.35\textwidth}
\centering
\includegraphics[height=4cm]{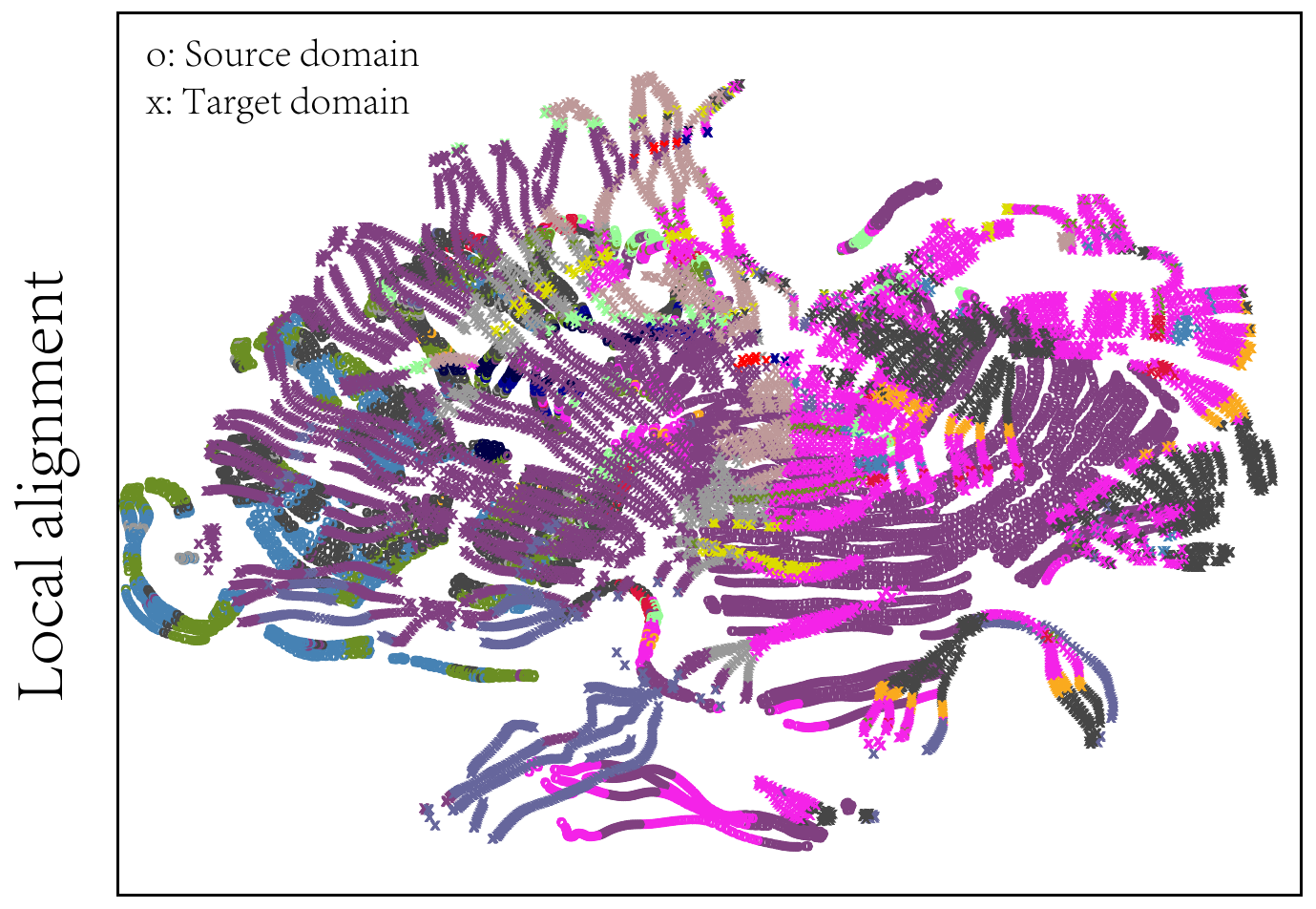}
\subcaption{Non-adapted}
\label{fig:tsne_NoAN_Class}
\end{minipage}
\begin{minipage}[b]{.32\linewidth}
\centering
\includegraphics[height=4cm]{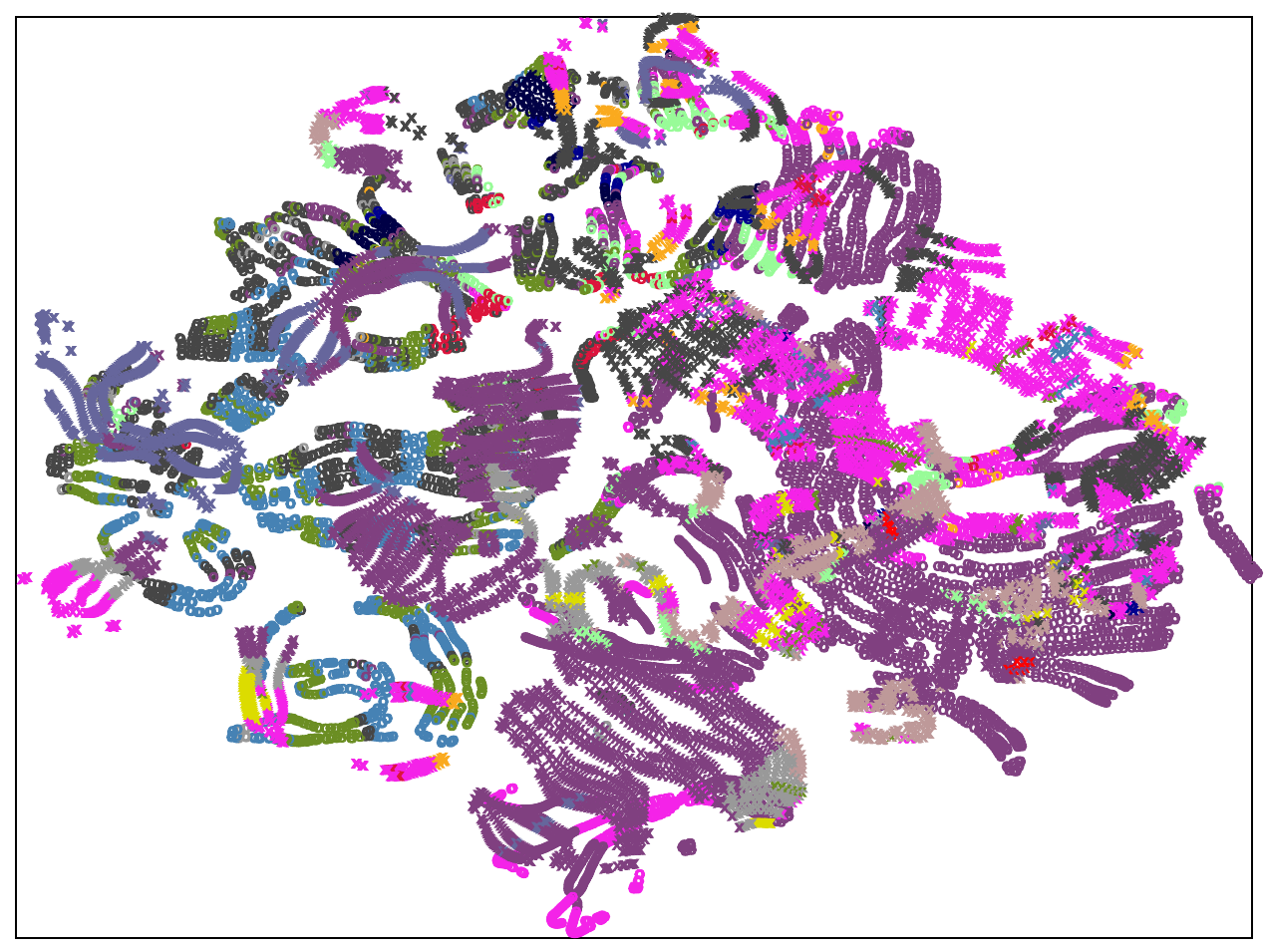}
\subcaption{IBAN}
\label{fig:tsne_IBAN_Class}
\end{minipage}
\begin{minipage}[b]{.32\linewidth}
\centering
\includegraphics[height=4cm]{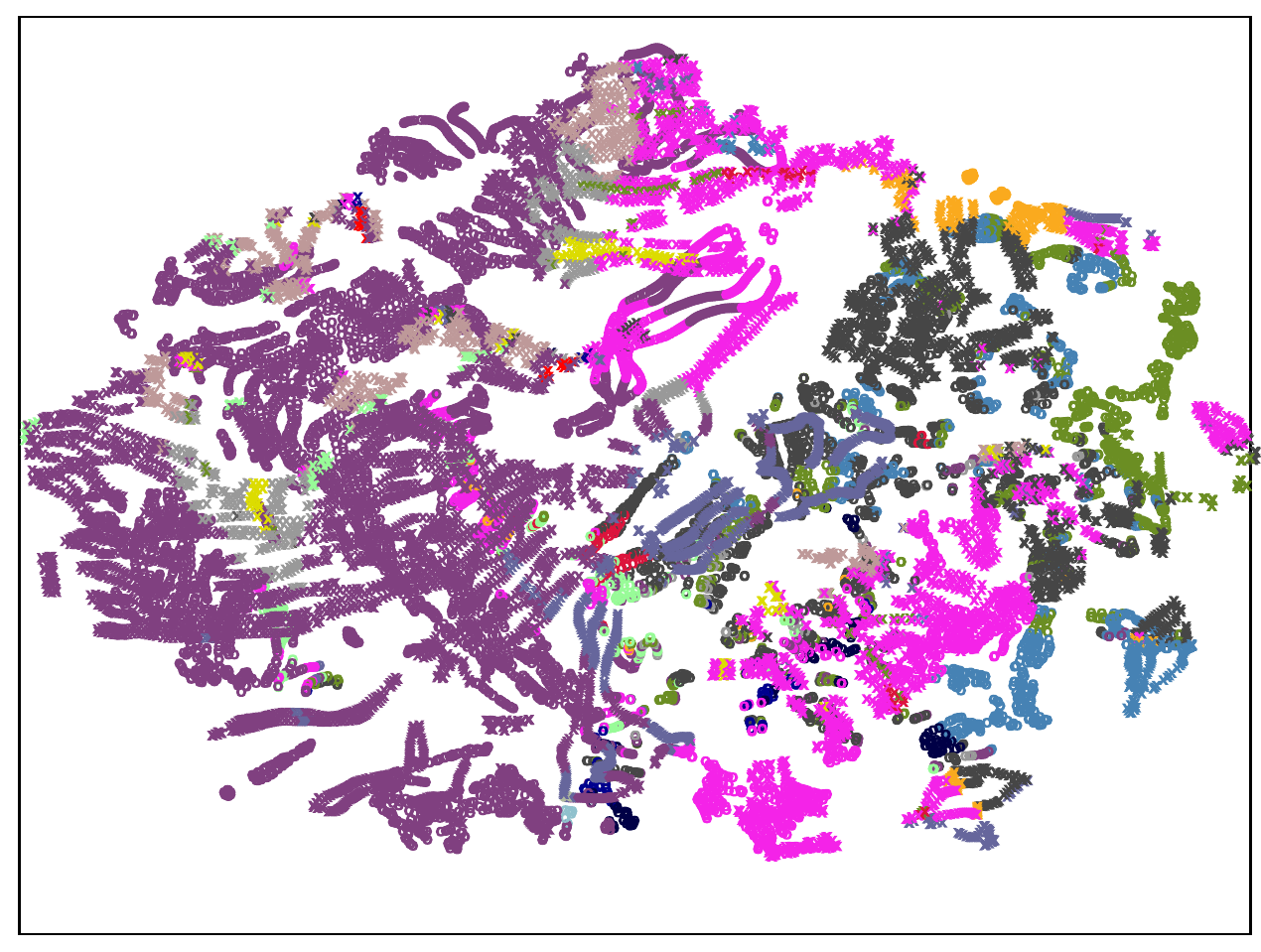}
\subcaption{SIBAN}
\label{fig:tsne_SIBAN_Class}
\end{minipage}
\caption{(Better zoom in.) We confirm the effects of SIBAN through a visualization of the learned representations $z_S$ \& $z_T$ using t-distributed stochastic neighbor embedding (t-SNE)~\cite{maaten2008tSNE}. Specifically, we show the results of Non-adapted model in (a)\&(d), IBAN in (b)\&(e) and SIBAN in (c)\&(f), respectively. In the first row, we label the t-SNE map by domains, where \textcolor{red}{red} denotes the source domain and \textcolor{blue}{blue} denotes the target domain. In the second row, we label the t-SNE map by different classes. The colors are consistent with the annotation maps.}
\label{fig:tsne}
\end{figure*}

In this section, we visualize the feature distributions in latent space aiming to confirm the effects of our method. To this end, we first select two similar images ($x_S$ and $x_T$) from source and target domain respectively and then map their high-dimensional latent features ($z_S$ and $z_T$) to a 2-D space with t-SNE~\cite{maaten2008tSNE} shown in Fig.~\ref{fig:tsne}. In the first row, we label the t-SNE maps by different domains in order to evaluate the marginal distribution alignment (\textbf{global alignment}) of the features between domains. While in the second row, we label the t-SNE maps by different semantic classes in order to evaluate the semantic consistency (\textbf{local alignment}) of the features between domains.

From the t-SNE maps, we can observe that the non-adaptive model can not yield well-aligned latent features, neither in global level (see Fig.~\ref{fig:tsne_NoAN_Domain}) or class level (see Fig.~\ref{fig:tsne_NoAN_Class}). These results demonstrate that the classifier trained on source data cannot be directly applied to target samples due to its limit generalization ability. For IBAN, the marginal distributions of the two domains are well aligned (see Fig.~\ref{fig:tsne_IBAN_Domain}), but some features from different semantic classes are mismatched (see Fig.~\ref{fig:tsne_IBAN_Class}). The reason lies in that the information constraint in IBAN, which enforces task-dependent features to a standard Gaussian distribution,  would wrongly compress the features from different classes too close to others and therefore make it hard to align them. Finally, we can see that SIBAN achieves good global and local feature alignment between domains (see Fig.~\ref{fig:tsne_SIBAN_Domain} \& Fig.~\ref{fig:tsne_SIBAN_Class}). The visualization of the latent feature distributions further explains why the SIBAN can achieve the leading results in feature-space adaptation.

\begin{figure*}[t]
    \centering
    \includegraphics[width=0.97\linewidth]{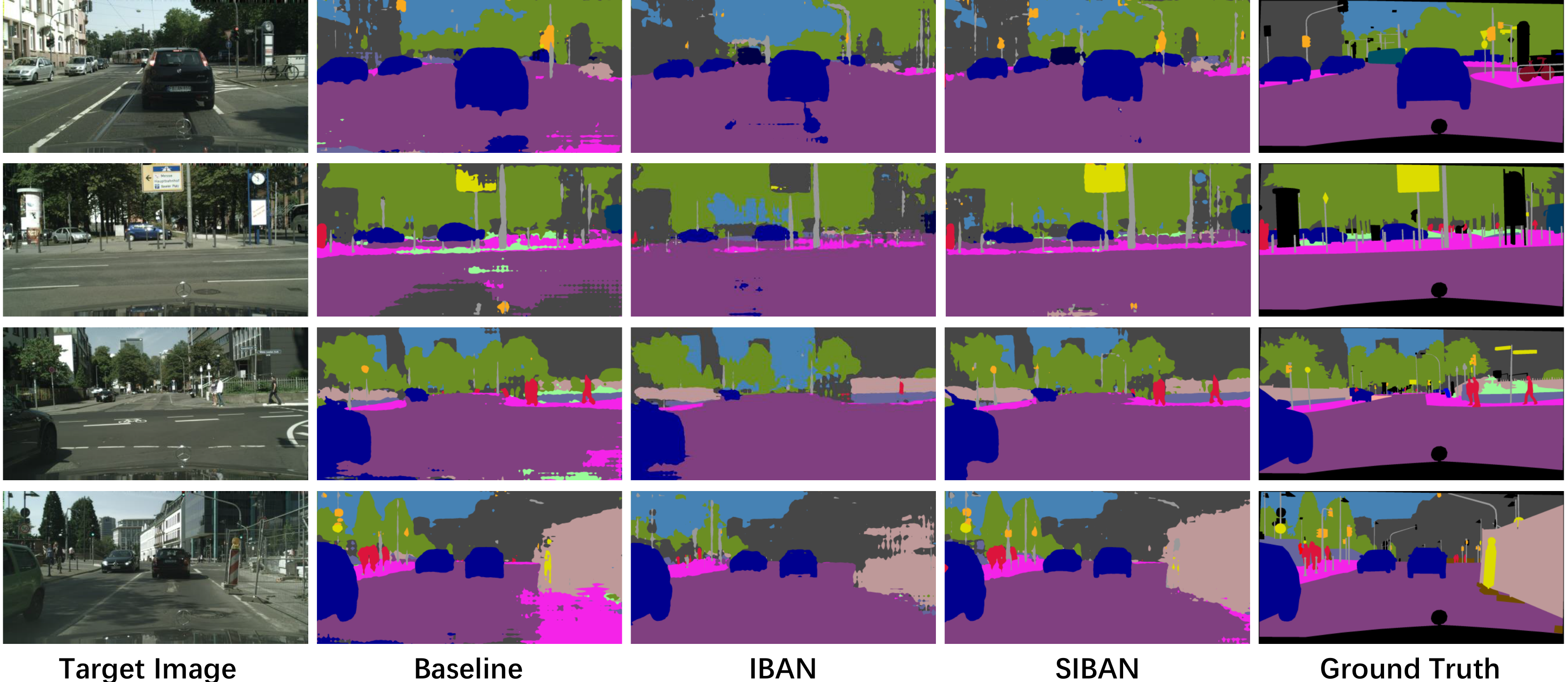}
    \caption{Qualitative results of the domain adaptive segmentation.}
    \label{fig:vis}
    \vspace{-0.45cm}
\end{figure*}

\section*{More Qualitative Results}
In Fig.~\ref{fig:vis}, we show more qualitative results from the baseline method, IBAN, and SIBAN respectively, followed by the ground truth label map.

\end{document}